\documentclass{article}
\usepackage{PRIMEarxiv}

\usepackage{amsmath}
\usepackage{algorithm}
\usepackage{algpseudocode}
\usepackage{array}
\usepackage[caption=false,font=normalsize,labelfont=sf,textfont=sf]{subfig}
\usepackage{textcomp}
\usepackage{stfloats}
\usepackage{pifont}  
\usepackage{url}
\usepackage{verbatim}
\usepackage{graphicx}
\usepackage{multirow}
\usepackage{tabularx}
\usepackage{color}
\usepackage{makecell}
\usepackage{diagbox}
\usepackage{enumerate}
\usepackage[table,xcdraw]{xcolor}
\usepackage{longtable}
\usepackage{threeparttable}
\usepackage[normalem]{ulem}
\usepackage{marginnote}
\usepackage[utf8]{inputenc} 
\usepackage[T1]{fontenc}    
\usepackage{hyperref}       
\usepackage{url}            
\usepackage{booktabs}       
\usepackage{amsfonts}       
\usepackage{nicefrac}       
\usepackage{microtype}      
\usepackage{lipsum}
\usepackage{fancyhdr}       
\usepackage{graphicx}       
\graphicspath{{media/}}     

\newtheorem{lemma}{\bf Lemma}
\useunder{\uline}{\ul}{}

\title{Search Multilayer Perceptron-Based Fusion for\\Efficient and Accurate Siamese Tracking
\thanks{\textbf{Accepted} by IEEE TCDS as a regular paper.
\textbf{Code:} \url{https://github.com/AmadeusSTQ/SEAT}.
\textsuperscript{${\dagger}$}Corresponding author.
}
}

\author{
  Tianqi Shen \\
  Institute of Mining Artificial Intelligence, Chinese Institute of Coal Science \\
  Department of Computer Science, City University of Hong Kong \\
  Beijing 100013, China \; / \; Hong Kong SAR \\
  \texttt{tianqshen5-c@my.cityu.edu.hk}
  \And
  Huakao Lin \\
  Image Processing Center, School of Astronautics, Beihang University \\
  Beijing 100191, China \\
  \texttt{sy2315212@buaa.edu.cn}
  \And
  Ning An$^{\dagger}$ \\
  Institute of Mining Artificial Intelligence, Chinese Institute of Coal Science \\
  State Key Laboratory of Intelligent Coal Mining and Strata Control \\
  Beijing 100013, China \\
  \texttt{ning.an.010@foxmail.com}
}

\begin{document}
\maketitle

\begin{abstract}
    Siamese visual trackers have recently advanced through increasingly sophisticated fusion mechanisms built on convolutional or Transformer architectures. However, both struggle to deliver pixel-level interactions efficiently on resource-constrained hardware, leading to a persistent accuracy–efficiency imbalance. Motivated by this limitation, we redesign the Siamese neck with a simple yet effective Multilayer Perception (MLP)-based fusion module that enables pixel-level interaction with minimal structural overhead. Nevertheless, naïvely stacking MLP blocks introduces a new challenge: computational cost can scale quadratically with channel width.
	To overcome this, we construct a hierarchical search space of carefully designed MLP modules and introduce a customized relaxation strategy that enables differentiable neural architecture search (DNAS) to decouple channel-width optimization from other architectural choices. This targeted decoupling automatically balances channel width and depth, yielding a low-complexity architecture.
	The resulting tracker achieves state-of-the-art accuracy–efficiency trade-offs. It ranks among the top performers on four general-purpose and three aerial tracking benchmarks, while maintaining real-time performance on both resource-constrained Graphics Processing Units (GPUs) and Neural Processing Units (NPUs).
\end{abstract}


\section{Introduction}
\begin{figure}[t]
	\centering
	\includegraphics[width=\linewidth]{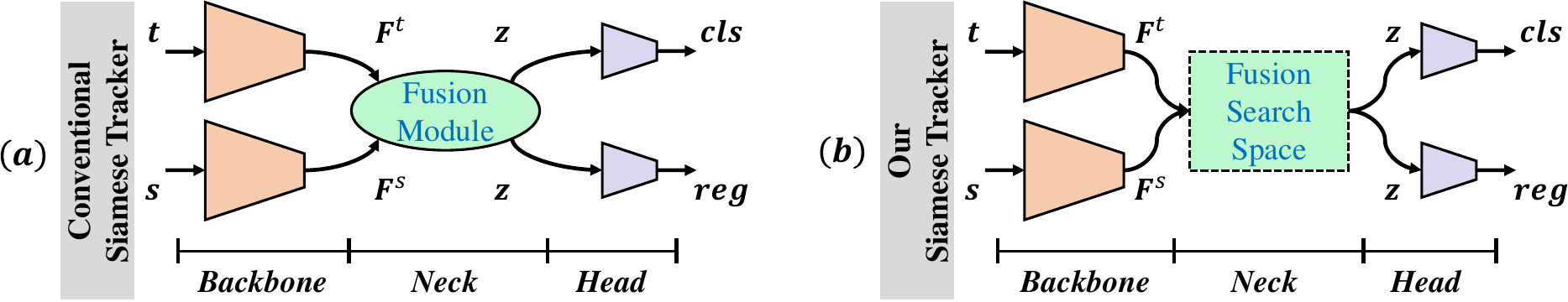}
	\caption{Comparison between the conventional pipeline and the proposed pipeline. In subfigure ($\boldsymbol{a}$), the template feature $\boldsymbol{F}^t$ and search feature $\boldsymbol{F}^s$ are processed using manually designed module(s). Alternatively, ($\boldsymbol{b}$) integrates the fusion architecture into a searchable space, enabling the discovery of a lightweight yet effective neck design.
    }
	\label{fig_pipe_compare}
\end{figure}

Visual object tracking is a fundamental task in computer vision, which aims to determine the location and scale of a target object within a video sequence \cite{zhang2023visual, 9913708}. Significant progress in this field \cite{liang2023global, borsuk2022fear, cao2021siamapn++, 9372336} has been driven by Siamese-based trackers. 
As illustrated in Figure~\ref{fig_pipe_compare}(a), these trackers typically consist of three core components \cite{bertinetto2016fully} that work together to achieve robust performance.
First, feature extraction is performed on both the template $\boldsymbol{t}$ and the search region $\boldsymbol{s}$ using a Siamese \textit{backbone}, which is often adapted from image recognition models \cite{zhang2019deeper, li2019siamrpn++, wang2023ldcnet, liu2020deep}. Second, the resulting feature maps (or tokens) $\boldsymbol{F}^t$ and $\boldsymbol{F}^s$ are fused in the \textit{neck} through correlation operations to generate one or more response maps $\boldsymbol{z}$. Finally, the state of the object, typically represented as a bounding box, is predicted by the tracking \textit{head}, which decodes $\boldsymbol{z}$ using classification ("cls") and regression ("reg") branches.

In this tracking paradigm, numerous studies \cite{guo2021graph, han2021learning, zhang2021learn, chen2021transformer} have highlighted that the correlation architecture, where feature fusion occurs, directly influences how the tracker utilizes prior information from the template to evaluate the object's new state in the search region. 
However, as fusion mechanisms become increasingly sophisticated, the challenge of \textbf{balancing accuracy and efficiency} has intensified, particularly on hardware-constrained platforms \cite{yan2021lighttrack, zhang2021learn, blatter2023efficient}. 
In other words, there remains a lack of modular fusion designs that support pixel-level interactions for accurate representation while maintaining high efficiency under limited computational resources. 

\begin{figure*}[h]
	\centering
	\includegraphics[width=\linewidth]{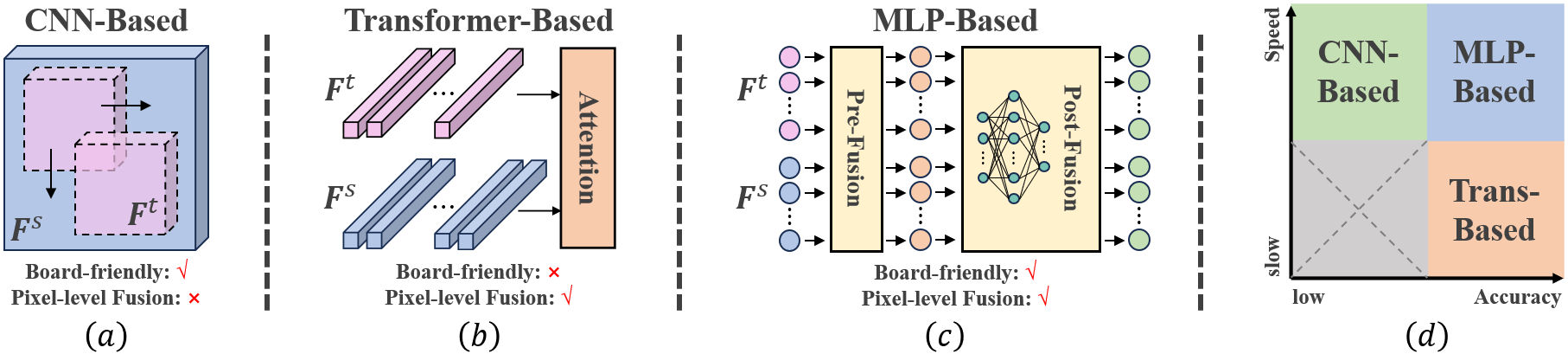}
	\caption{($\boldsymbol{a}$) CNN-based fusion uses template feature maps $\boldsymbol{F}^t$ as kernels sliding over search feature maps $\boldsymbol{F}^s$, performing patch-level fusion efficiently on hardware. ($\boldsymbol{b}$) Transformer-based fusion applies cross-attention between template tokens $\boldsymbol{F}^t$ and search tokens $\boldsymbol{F}^s$, achieving pixel-level fusion but with slower hardware execution. ($\boldsymbol{c}$) Our MLP-based fusion first performs coarse fusion on $\boldsymbol{F}^t$ and $\boldsymbol{F}^s$, followed by refinement using Wave-MLP blocks, resulting in hardware-efficient pixel-level fusion. ($\boldsymbol{d}$) The MLP-based approach yields a better accuracy–speed trade-off than its CNN and Transformer counterparts.}
	\label{fig_abcd}
\end{figure*}

From the accuracy perspective, precise feature fusion in the correlation stage is essential, as the performance of the tracking head heavily depends on accurate similarity estimation. Therefore, the fusion module must operate at the pixel level to ensure that each pixel in the response map reflects the similarity between features from the template and the search region. As shown in Figure~\ref{fig_abcd}, existing methods have progressed from patch-level convolutional neural network (CNN)-based fusion to pixel-level Transformer-based fusion \cite{zhang2021learn}. However, this shift has led to a significant increase in computational cost \cite{chen2021transformer}.

To address this issue, we leverage the fully connected nature of the multilayer perceptron (MLP) to enable efficient pixel-level fusion. Specifically, we propose two novel modules. The \textit{Coarse Fusion MLP (CFM)} performs early integration of template and search features in a simple yet effective manner, analogous to attention mechanisms. The \textit{Refine Fusion MLP (RFM)} employs Wave-MLP blocks \cite{tang2022image} to further enhance and refine the fused features. Together, these modules form a progressive fusion framework that transitions from coarse to fine levels of detail.

From the efficiency perspective, adopting an MLP-based fusion module presents \textbf{a new challenge}: MLP blocks, particularly Wave-MLP, exhibit quadratic computational complexity w.r.t the channel dimension (see Lemma \ref{PATM}). This complexity makes it difficult to enhance representational capacity without compromising efficiency on resource-constrained hardware, motivating the use of \textit{Neural Architecture Search (NAS)} for automated design.
However, while prior lightweight trackers have employed NAS to optimize the backbone and prediction head \cite{yan2021lighttrack, borsuk2022fear, blatter2023efficient}, the correlation architecture in the neck (Figure~\ref{fig_pipe_compare}(b)) has received comparatively little attention. 

To address this gap, we integrate the CFM and RFM modules into a hierarchical \textit{MLP-based Correlation Architecture Space (MCAS)}, complemented by a \textit{Harmony-Relaxation} strategy. Unlike conventional flat search spaces, MCAS organizes Wave-MLP variants that share the same channel count but differ in other hyperparameters into distinct \textit{Harmonization blocks}, each defined at a specific channel dimension. This hierarchical design decouples the optimization of channel width across Harmonization blocks from the fine-grained tuning of other architectural parameters within each block. 

In summary, our methods \textbf{\textit{S}}earch for and yield, to the best of our knowledge, \textbf{the first} MLP-based fusion architecture explicitly designed for \textbf{\textit{E}}fficient and \textbf{\textit{A}}ccurate Siamese \textbf{\textit{T}}racking under resource-constrained settings. We refer to the resulting tracker as \textbf{\textit{SEAT}}. Our key contributions are as follows:

\begin{itemize}
	\item We propose a simple yet effective MLP-based coarse-to-fine fusion framework, comprising the CFM and RFM modules, which enables efficient pixel-level feature interaction and improves Siamese tracker accuracy.
	\item We introduce a hierarchical MCAS with a Harmony-Relaxation strategy that supports DNAS and decouples channel-width optimization from other architectural hyperparameters, enhancing Siamese tracker efficiency.
	\item We develop two lightweight tracker variants: SEAT\_LT for resource-limited Graphics Processing Units (GPUs) and SEAT\_AL for resource-constrained Neural-network Processing Units (NPUs).
\end{itemize}

The remainder of the paper is organized as follows. Section \ref{section:related_work} reviews three pertinent research areas that contextualize our contributions. Section \ref{section:methodology} presents the technical details (based on theoretical analysis) of our hierarchical MCAS search space, together with the Harmony-Relaxation strategy. Section \ref{section:experiments} provides comprehensive evaluations against state-of-the-art (SOTA) methods across multiple benchmarks. Finally, Section \ref{section:conclusion} summarizes the findings and outlines potential directions for future work.

\section{Related Work}
\label{section:related_work}
In this section, we review the current landscape of three fields that this work intersects, highlighting the key developments and advancements in each field. We also demonstrate how our work fills the gap at the intersection of them.

\subsection{Feature Fusion in Siamese-Based Trackers}

Early Siamese trackers, such as SiamFC \cite{bertinetto2016fully}, adopted cross-correlation by treating template features as convolutional kernels applied to the search features. This CNN-based fusion approach is efficient and compatible with variable-size inputs. However, it operates at the patch level, limiting its ability to capture fine-grained, pixel-level similarity.
Subsequent works \cite{chen2020siamese, guo2020siamcar, han2021learning, zhang2021learn} explored deeper or asymmetric correlation modules and combined multiple fusion operators. Despite these advancements, they remained constrained by the coarse granularity inherent to convolutional operations.
To improve alignment precision, recent methods have transitioned to Transformer-based fusion. These models employ cross-attention to directly model pixel-to-pixel relationships between the template and search regions \cite{chen2021transformer, gao2022aiatrack, han2022deep}. Some even unify the backbone and fusion components within a single Transformer network \cite{cui2022mixformer, ye2022joint, han2024two}. While these designs offer high accuracy, they also introduce significant computational overhead, making them less suitable for real-time edge deployments.
As illustrated in Figure~\ref{fig_abcd} and Figure~\ref{fig:pipeline}, we propose an MLP-based fusion framework that performs pixel-level integration while substantially reducing computational complexity through the design of the CFM and RFM modules. 

\subsection{Vision MLP}
\label{subsection:vision_mlp}

Recent Vision MLPs have emerged as a powerful paradigm, demonstrating remarkable performance in various tasks \cite{liu2022we}.
MLP-Mixer \cite{tolstikhin2021mlp} introduces a simple architecture composed of two MLP layers to extract features both within and across image patches. Subsequent works \cite{tang2022sparse, guo2022hire} extended this design by reducing computational cost through axial aggregation and by hierarchically rearranging tokens to better capture both local and global information. Wave-MLP \cite{tang2022image} further advances the field by representing each token as a wave function with amplitude and phase, allowing it to model content variations across different images more effectively.
Despite this progress, Vision MLPs have been scarcely explored in visual tracking. The sole instance, MLPT~\cite{chan2022mlpt}, appends an MLP-Mixer-style block after a CNN-based cross-correlation module rather than redesigning the whole fusion mechanism itself. Such limited adoption stems from two inherent limitations that impede direct application in Siamese tracking: (1) most Vision MLPs are built for single-stream inputs and do not naturally fuse two heterogeneous feature sources
($\boldsymbol{F}^t$ and $\boldsymbol{F}^s$);
and (2) although structurally simple, stacking MLP blocks, particularly Wave-MLP, induces quadratic growth in computational complexity w.r.t channel number, as shown in Lemma \ref{PATM}.
Our approach tackles these challenges by performing pre-fusion via CFM to handle dual-stream inputs (addressing limitation (1)), and by organizing CFM and RFM within a hierarchical MCAS that decouples channel width from other hyperparameters, thereby improving efficiency via DNAS (addressing limitation (2)).

\subsection{Neural Architecture Search for Tracking}

\begin{figure*}[h]
	\centering
	\includegraphics[width=\linewidth]{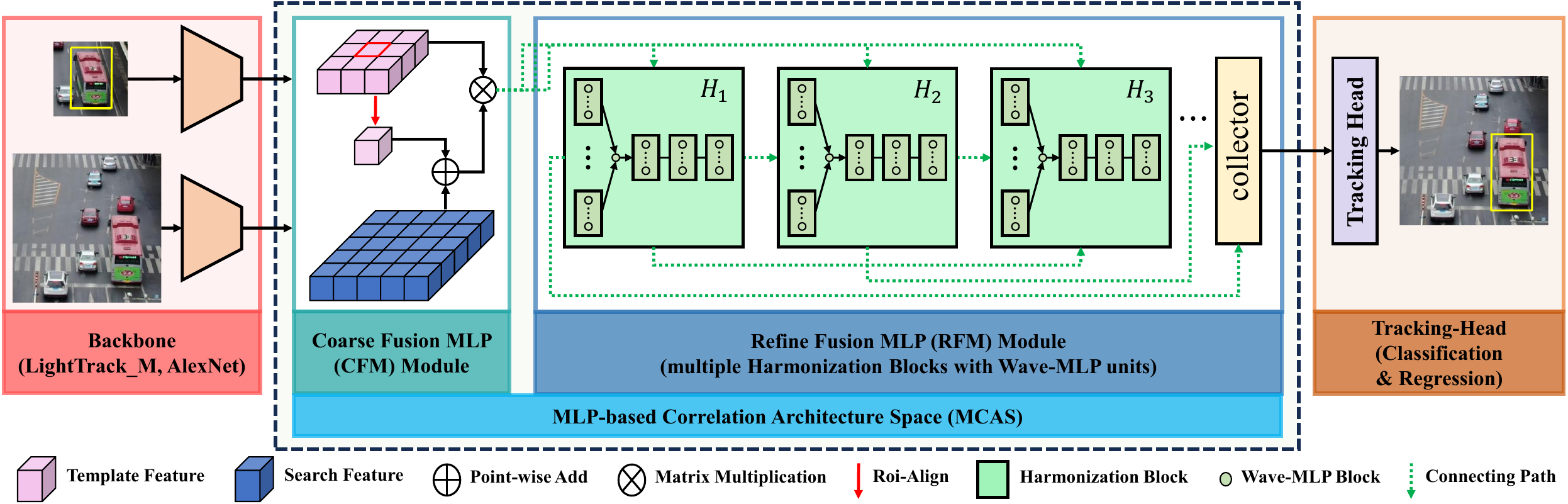}
	\caption{Overview of our proposed tracker and MCAS. 
		\textbf{i.} Versatile architecture with two backbones: \cite{yan2021lighttrack} for resource-constrained GPU platforms and \cite{krizhevsky2017imagenet} for resource-limited NPU deployment.
		\textbf{ii.} The primary MLP-based components in the MCAS include the Coarse Fusion MLP (CFM) Module and the Refine Fusion MLP (RFM) Module. CFM performs coarse fusion to generate the heatmap, while RFM refines it to produce the response map.
		\textbf{iii.} Tracking-head network transforms the response map into classification and regression response maps, representing the predictive classification and bounding box of the target.}
	\label{fig:pipeline}
\end{figure*}

Neural Architecture Search (NAS) has been widely adopted to automate network design \cite{zoph2017neural}. Early methods based on reinforcement learning (RL) or evolutionary algorithms (EA) are often computationally expensive and limited to coarse design choices \cite{pham2018efficient}. Differentiable NAS (DNAS) approaches such as DARTS \cite{liu2018darts} and FBNet \cite{wu2019fbnet} improve efficiency mainly by employing weight-sharing strategies to assess candidate architectures \cite{song2024efficient}.
In visual tracking, LightTrack \cite{yan2021lighttrack} uses EA-based NAS to search for lightweight CNN backbones, while FEAR \cite{borsuk2022fear} and EfficientTrack \cite{blatter2023efficient} directly adopt architectures searched for image classification. UAV-oriented trackers such as SiamAPN++ \cite{cao2021siamapn++}, HIFT \cite{cao2021hift}, and TCTrack \cite{cao2022tctrack} rely on shallow, manually designed CNNs (e.g., AlexNet \cite{krizhevsky2017imagenet}) to prioritize speed.
However, prior work has largely overlooked NAS-driven design for feature fusion, particularly within the neck—the critical component for object localization in visual tracking. Moreover, vanilla DARTS cannot naturally accommodate dual-input fusion nor support decoupled optimization of channel width (see Section \ref{subsection:vision_mlp} limitation (2)). As shown in Figure~\ref{fig:pipeline}, we address these gaps by introducing MCAS with Harmony-Relaxation. Specifically, Wave-MLP blocks that share the same channel count but differ in other hyperparameters (e.g., kernel size, MLP expansion ratio) are grouped into Harmonization blocks. Along the inter-block "highway" (in green dashed arrow), we apply a relaxation parameter $\beta$ to optimize channel width, while within each block $H_i$ we employ a relaxation parameter $\alpha$ to select among candidate operations. This yields, to our knowledge, the first DNAS framework explicitly tailored to designing lightweight fusion modules for Siamese trackers.

\section{Methodology}
\label{section:methodology}
In this section, we first provide some preliminary background and then introduce our novel MLP-based Correlation Architecture Space (MCAS) method, highlighting its innovative aspects and contributions to the field.

\subsection{Theoretical Foundations and Extensions}
\subsubsection{Wave-MLP Block}
\label{subsubsection:computational_analysis}
The Wave-MLP block achieves token mixing through Phase-Aware Token Mixing (PATM) \cite{tang2022image}, which models each token as a wave with learned amplitude and phase. The fully connected layers in PATM establish global dependencies, providing a natural pixel-level fusion to enhance tracking accuracy.

Given an arbitrary feature map $\boldsymbol{z}\in\mathbb{R}^{H_sW_s\times C'}$, the Wave-MLP block transforms each token $\boldsymbol{z}_\phi\in\mathbb{R}^{1\times C'}$ through the following equations:
\begin{equation}
	\boldsymbol{h}_\phi = \boldsymbol{z}_\phi\boldsymbol{W}^h 
	\in \mathbb{R}^{C'}, 
	\quad
	\boldsymbol{\theta}_\phi = \boldsymbol{z}_\phi\boldsymbol{W}^{\theta} 
	\in \mathbb{R}^{C'},
	\nonumber
\end{equation}
where $\boldsymbol{h}_\phi$ and $\boldsymbol{\theta}_\phi$ represent the amplitude and phase of the $\phi$-th token, respectively, and $\boldsymbol{W}^h$ and $\boldsymbol{W}^{\theta}$ are trainable parameters of size $C' \times C'$.

Token mixing is then performed using Euler's formula \cite{euler1797introductio}, projecting the complex wave state to real values:
\begin{equation}
	\boldsymbol{o}_{\phi'} = \sum_{\phi=1}^{H_sW_s} {
		(\boldsymbol{h}_\phi\odot 
		{\rm cos}\boldsymbol{\theta}_\phi) 
		\boldsymbol{W}^{cos}_{\phi'\phi} + 
		(\boldsymbol{h}_\phi\odot 
		{\rm sin}\boldsymbol{\theta}_\phi) 
		\boldsymbol{W}^{sin}_{\phi'\phi}
	},
	\tag{\text{PATM mechanism in Wave-MLP}}
\end{equation}
where $\odot$ denotes element-wise multiplication, $\boldsymbol{o}_{\phi'}$ is the $\phi'$-th token after mixing, and $\boldsymbol{W}^{cos}_{\phi'\phi}$ and $\boldsymbol{W}^{sin}_{\phi'\phi}$ are trainable parameters of size $C' \times C'$.

However, naively stacking Wave-MLP blocks can lead to increased computational costs, as shown in our Lemma \ref{PATM} (the proof can be found in Appendix~\ref{append:computational_analysis_of_wavemlp}).
\begin{lemma} \label{PATM}
	Let the channel number of a PATM module be $C'$. Then, the number of FLoating Point Operations (FLOPs) of this module is bounded by $O({C'}^2)$.
\end{lemma}

According to Lemma \ref{PATM}, the FLOPs of Wave-MLP blocks increase quadratically with the channel number $C'$. Therefore, although a larger channel number $C'$ can lead to a more abundant feature expression, the channel number $C'$ should be discretely set to maintain a lightweight tracker.

\subsubsection{DARTS}
DARTS~\cite{liu2018darts} formulates NAS by relaxing the discrete search space $\mathcal{A}$ into a continuous one, referred to as a \textit{supernet}. In this context, \textit{relaxation} means that for each edge in the supernet, the candidate operations $\mathcal{O}$ are weighted using a softmax function over architecture parameters $\alpha_{\mathcal{O}} := \{\alpha_{o} \mid o \in \mathcal{O}\}$:
\begin{equation}
    \text{output} = \sum_{o \in \mathcal{O}} \frac{\exp(\alpha_o)}{\sum_{o' \in \mathcal{O}} \exp(\alpha_{o'})} o(\text{input}).
    \label{eq:relaxation}
\end{equation}
Here, $\alpha_{o}$ is the architecture parameter assigned to operation $o$. Each operator (e.g., a Wave-MLP block) has its own trainable weights $w$, which collectively form the set of all trainable weights $W$ in the supernet. This relaxation allows discrete operations to be embedded into the supernet $\mathcal{N}(\alpha, W)$, making it continuously differentiable.
This formulation~\eqref{eq:relaxation} enables gradient-based optimization through bi-level training. In the inner loop, $W$ is optimized to minimize the training loss $\mathcal{L}_{\text{train}}$ on the training dataset. In the outer loop, $\alpha$ is optimized to maximize the validation accuracy $\text{Acc}_{\text{val}}$ on the validation set, leveraging continuous differentiability:
\begin{equation}
    \begin{aligned}
        \alpha^* &= \arg \max_{\alpha\in\mathcal{A}} \text{Acc}_{\text{val}}\left(\mathcal{N}(\alpha, W^*(\alpha))\right), \\
        \text{s.t.} \quad W^*(\alpha) &= \arg \min_W \mathcal{L}_{\text{train}}(\mathcal{N}(\alpha, W)).
    \end{aligned}
    \nonumber
\end{equation}
Here, $\alpha$ is formed by grouping all $\alpha_{\mathcal{O}}$'s values across all sets of operations $\mathcal{O}$, i.e., $\alpha:=\bigcup_\mathcal{O}\alpha_{\mathcal{O}}$. In summary, the inner loop optimizes the supernet weights $W$ while keeping $\alpha$ fixed, and the outer loop updates $\alpha$ while fixing $W$.

This approach enables efficient architecture search by eliminating the need to train each candidate from scratch \cite{li2022abcp}, a strategy known as weight sharing \cite{song2024efficient}. Importantly, DARTS treats all architecture parameters $\alpha$ uniformly under a single optimizer. Although it can optimize discrete parameters such as channel numbers using certain techniques (e.g., those employed in FBNetV2 \cite{wan2020fbnetv2}), it does so independently and not in a truly decoupled manner. This limitation makes it suboptimal for MLP-based fusion modules, where channel count is the dominant factor in computational complexity (see Lemma~\ref{PATM}).
To address this limitation, we propose a hierarchical MCAS framework that adapts DARTS to MLP-based search spaces. It introduces a harmonization block to enable channel-aware DNAS by decoupling the choice of feature channel numbers (controlled by a new set of architectural-parameters $\beta$) from other architectural factors such as kernel size and expansion ratio (controlled by the typical set of architectural-parameters $\alpha$). The two parameter sets are optimized in a fully decoupled manner using separate optimizers. Further details are provided in the following subsections.

\subsection{MLP-based Correlation Architecture Space (MCAS)}
\label{subsection:mcas}

\subsubsection{Coarse Fusion MLP Module (CFM)}
\label{subsection:Coarse_Fusion_MLP_Module}
As illustrated in Figure~\ref{fig:pipeline}, the CFM Module which serves as the first fixed entity in the MCAS, is responsible for initially fusing $\boldsymbol{F}^t$ and $\boldsymbol{F}^s$. 
The pseudocode is provided in Algorithm~\ref{alg:CFM}, where $+_b$ and $S$ denote broadcast summation and tensor reshaping, respectively.
In this algorithm, Operation 1 isolates the region of interest (ROI) for subsequent enhancement. Pixel-level fusion occurs in Operations 2 and 3, while Operation 4 maps the channel features into a more distinct space, facilitating processing by the subsequent modules. Although the algorithm's steps are straightforward, Lemma \ref{lemma:CFM} demonstrates that CFM can achieve effects akin to an attention mechanism (see Appendix~\ref{append:similarity_measurement_of_cfm}).

\begin{algorithm}
	\caption{Coarse Fusion MLP (CFM) Module}
	\label{alg:CFM}
	\begin{algorithmic}[1]
		\Require $\boldsymbol{F}^t \in \mathbb{R}^{H_t \times W_t \times C}$: Template feature map
		\Require $\boldsymbol{F}^s \in \mathbb{R}^{H_s \times W_s \times C}$: Search feature map
		\Require $bbox$: Bounding box of the target object in the $\boldsymbol{t}$
		\Ensure $\boldsymbol{z} \in \mathbb{R}^{H_s W_s \times C'}$: Coarsely fused heat map
		\Statex \textbf{Operation 1: ROI Align}
		\State $\boldsymbol{r} \leftarrow \text{ROIAlign}(\boldsymbol{F}^t, bbox) \in \mathbb{R}^{1 \times 1 \times C}$
		\Comment{Extract target region features from the template}
		\Statex \textbf{Operation 2: Point-wise Addition}
		\State $\boldsymbol{r}' \leftarrow \text{Duplicate}(\boldsymbol{r}, H_s \times W_s) \in \mathbb{R}^{H_s \times W_s \times C}$
		\State $\boldsymbol{x} \leftarrow \boldsymbol{r}' +_b \boldsymbol{F}^s \in \mathbb{R}^{H_s \times W_s \times C}$
		\Comment{Broadcast and sum with the search region features}
		\Statex \textbf{Operation 3: Matrix Multiplication}
		\State $\boldsymbol{x}' \leftarrow S(\boldsymbol{x}) \in \mathbb{R}^{H_s W_s \times C},\quad \boldsymbol{F}^{t'} \leftarrow S(\boldsymbol{F}^t) \in \mathbb{R}^{C \times H_t W_t}$
		\State $\boldsymbol{y} \leftarrow \text{MatMul}(\boldsymbol{x}', \boldsymbol{F}^{t'}) \in \mathbb{R}^{H_s W_s \times H_t W_t}$
		\Comment{Multiply reshaped feature maps}
		\Statex \textbf{Operation 4: Channel Transformation}
		\State $\boldsymbol{z} \leftarrow \boldsymbol{y} \boldsymbol{W} + \boldsymbol{b} \in \mathbb{R}^{H_s W_s \times C'}$
		\Comment{Transform to match input channel numbers of RFM}
	\end{algorithmic}
\end{algorithm}

\begin{lemma} \label{lemma:CFM}
	The output $\boldsymbol{z} \in \mathbb{R}^{H_s W_s \times C'}$ from the CFM can be computed using the following equation, which defines the element $z_{\phi\zeta}$ located at row $\phi$ and column $\zeta$:
	\begin{equation} 
		\sum_{\psi=1}^C 
		{
			F^s_{\phi\psi}
			(
			\sum_{\varphi=1}^{H_tW_t} {F^t_{\psi\varphi}w_{\varphi\zeta}}
			)
		} 
		+ 
		\sum_{\psi=1}^C 
		{
			r_\psi
			(
			\sum_{\varphi=1}^{H_tW_t} {F^t_{\psi\varphi}w_{\varphi\zeta}}
			)
		} 
		+ b_{\phi\zeta},
		\label{eq:CFM}
	\end{equation}
    where $F^s_{\phi\psi} + r_\psi$ corresponds to the element at row $\phi$ and column $\psi$ of $S(\boldsymbol{x})$, and $F^t_{\psi\varphi}$ is the corresponding element in $S(\boldsymbol{F^t})$. The terms $w_{\varphi\zeta}$ and $b_{\phi\zeta}$ are elements of the trainable weight matrix $\boldsymbol{W}$ and bias vector $\boldsymbol{b}$, as defined in Operation 4 of Algorithm~\ref{alg:CFM}.
\end{lemma}

The latter term of Equation~\eqref{eq:CFM} aligns the target region's features with those of the template using trainable parameters. This process enhances the representation of the object region while comparatively diminishing that of the background, similar to the self-attention mechanism in the ECA module of TransT~\cite{chen2021transformer}. In contrast, the former term of Equation~\eqref{eq:CFM} computes the pixel-level similarity between the template and search feature maps using trainable parameters, resembling the cross-attention mechanism of TransT's CFA module.

By generating the heat map $\boldsymbol{z}$ through CFM, an initial fusion of $\boldsymbol{F}^t$ and $\boldsymbol{F}^s$ is achieved. This allows the subsequent Wave-MLP blocks in RFM to process $\boldsymbol{z}$ directly, facilitating the fusion of more distinct feature tokens at the pixel level within a compact and discriminative feature space, ultimately improving the accuracy of our tracker. Therefore, the CFM module addresses challenge (1) from Section \ref{subsection:vision_mlp}.

\subsubsection{Refine Fusion MLP Module (RFM)}
\label{subsection:Refine_Fusion_MLP_Module}

\begin{figure}[h]
	\centering
	\includegraphics[width=10cm]{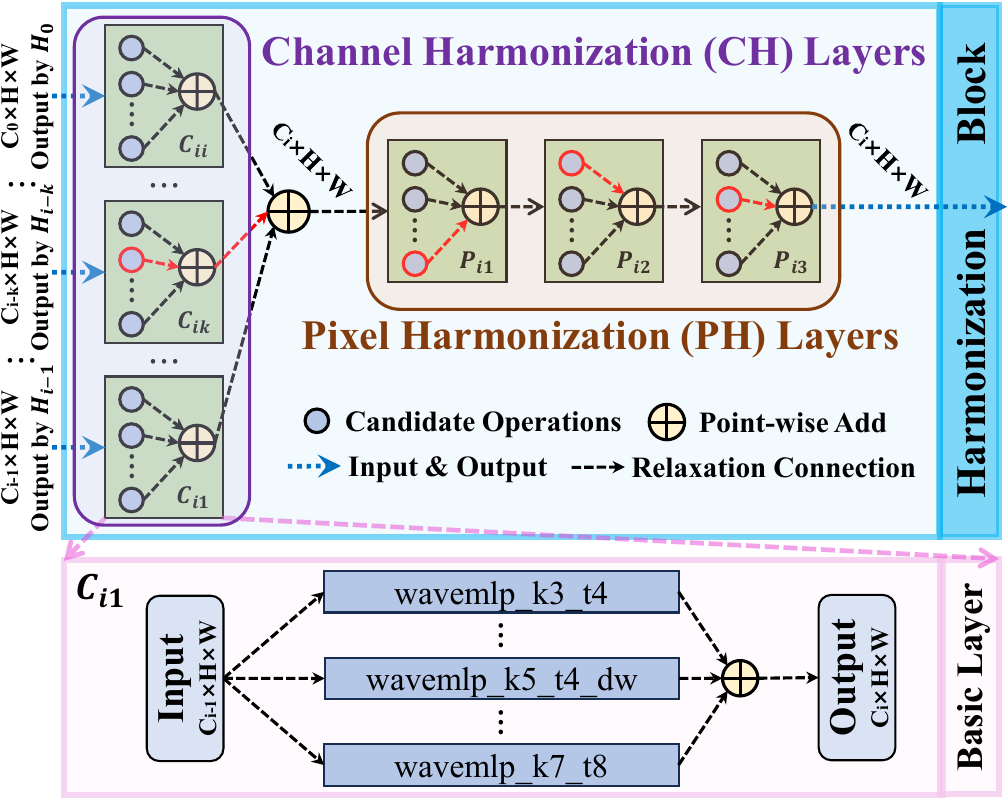}
	\caption{
		Structure of the harmonization block $H_i$ and its basic layer $C_{i1}$. 
		\textbf{Top:} Basic layers are categorized into CH layers ($C_{i1},\cdots,C_{ik},\cdots,C_{ii}$ with $1\leq k\leq i$) and PH layers ($P_{i1},P_{i2},P_{i3}$). 
		\textbf{Bottom:} $C_{i1}$ consists of 12 Wave-MLP blocks with different configurations. 
		"wavemlp\_k5\_t4\_dw" refers to a Wave-MLP block where the PATM module uses a $5{\times}5$ convolution kernel ("k5"), the MLP expansion ratio is 4 ("t4"), and depthwise separable convolution is applied ("dw"). \textit{Each CH or PH layer is a basic layer.}
	}
	\label{fig:harmoni}
\end{figure}

As shown in Figure~\ref{fig:pipeline}, the RFM module derived from the MCAS, serves as a flexible component responsible for enhancing and fusing the features $\boldsymbol{F}^t$ and $\boldsymbol{F}^s$ through a lightweight configuration and connection scheme. The primary goal here is to simplify the complex search space (excluding CFM) by grouping Wave-MLP blocks with diverse configurations into a novel structure, referred to as the harmonization block, as illustrated in Figure~\ref{fig:harmoni}.

\paragraph{Harmonization Block}
MCAS employs a set of distinct harmonization blocks, denoted as $H_i$, to search for an appropriate RFM. Each block aggregates feature maps from preceding layers and transmits the fused results to subsequent ones. This cascading structure introduces multiple pathways within the supernet, enabling more flexible and effective feature fusion.
We employ one CFM and three harmonization blocks, denoted as $H_0$, $H_1$, $H_2$, and $H_3$.
Regardless of the backbone architecture, $H_0$ transforms the input feature map's channel count to 256 and forwards the result to the subsequent blocks, where the spatial resolution of the feature map is not altered.
The specific interconnections among these blocks are depicted by the green dashed lines in Figure~\ref{fig:pipeline}.

\begin{table}[h]
	\caption{CH and PH Configurations of Each Harmonization Block}
	\label{tab:ch_ph}
	\centering
	\resizebox{0.58\linewidth}{!} 
	{
		\begin{threeparttable} 
			\begin{tabular}{ccccc}
				\hline
				\multirow{2}{*}{$H_1$} & 
				CH & 
				$C_{11} (256,256)$ & 
				- & 
				- \\ & 
				PH & 
				$P_{11} (256,256)$ & 
				$P_{12} (256,256)$ & 
				$P_{13} (256,256)$ 
				\\ \hline
				\multirow{2}{*}{$H_2$} & 
				CH & 
				$C_{22} (256,320)$ &
				$C_{21} (256,320)$ &
				- \\ & 
				PH & 
				$P_{21} (320,320)$ & 
				$P_{22} (320,320)$ & 
				$P_{23} (320,320)$ 
				\\ \hline
				\multirow{2}{*}{$H_3$} & 
				CH & 
				$C_{33} (256,384)$ &
				$C_{32} (256,384)$ &
				$C_{31} (320,384)$
				\\ & 
				PH & 
				$P_{31} (384,384)$ & 
				$P_{32} (384,384)$ & 
				$P_{33} (384,384)$ \\ \hline
			\end{tabular}
			\begin{tablenotes}  
				\footnotesize          
				\item[1] $C_{31}(320,384)$ indicates that the input channel count is 320 and the output dimension is 384. The same notation applies to other entries.
			\end{tablenotes}       
		\end{threeparttable} 
	}
\end{table}

\paragraph{CH Layer and PH Layer}
Each harmonization block consists of two primary components: Channel Harmonization (CH) layers and Pixel Harmonization (PH) layers. The CH layers comprise several parallel basic layers that receive feature maps with varying channel dimensions from earlier blocks. These inputs are unified using convolutional layers to align the channel dimensions. The processed feature maps are then forwarded to the PH layers, where further enhancement and fusion are performed at the pixel level using additional basic layers. 
Let $C_{ik}$ and $P_{ij}$ denote the $k$-th and $j$-th basic layers in the CH and PH layers of $H_i$, respectively. The basic layer $C_{ik}$ is responsible for processing the output from $H_{i-k}$.
The configurations of $C_{ik}$ and $P_{ij}$ in each $H_i$ are listed in Table~\ref{tab:ch_ph}, in which 256, 320, and 384 are selected as the fixed candidate channel dimensions based on prior knowledge derived from recent Transformer-based trackers (e.g., \cite{zhang2023siamst, zheng2024odtrack, han2024two}).

\paragraph{Wave-MLP Variants in Basic Layer}
Each basic layer in harmonization block is constructed by grouping a set of 12 Wave-MLP variants (as listed in Table \ref{tab:wavemlp_variant}) along with a skip connection. If the skip connection is selected during architecture search, the corresponding basic layer is effectively bypassed.
Since CH layers are responsible for aligning the channel dimensions across different feature streams, the skip connection is excluded from the candidates. In contrast, PH layers perform pixel-level refinement without altering channel dimensions, and therefore include the skip connection as a valid candidate.
In CH layers, candidate Wave-MLP blocks are slightly modified to support channel transformation. Specifically, while original Wave-MLP designs maintain input-output channel consistency, we adapt the first convolutional layer to project the input into the desired output channel dimension. 
The transformed features are then passed to the PATM module for spatial fusion. For brevity, we do not include a detailed architectural diagram of the Wave-MLP block in this paper.

\begin{table}[h]
	\caption{Configurations of Different Wave-MLP Variants}
	\label{tab:wavemlp_variant}
	\centering
	\resizebox{0.54\linewidth}{!} 
	{
		\begin{threeparttable} 
			\begin{tabular}{cccc}
				\hline
				Type       & Expansion Ratio & Kernel Size & Depthwise \\ \hline
				k3\_t4     & 4               & 3           
				& \ding{55}          \\
				k3\_t4\_dw & 4               & 3           
				& \checkmark          \\
				k3\_t8     & 8               & 3           
				& \ding{55}          \\
				k3\_t8\_dw & 8               & 3           
				& \checkmark          \\
				k5\_t4     & 4               & 5           
				& \ding{55}          \\
				k5\_t4\_dw & 4               & 5           
				& \checkmark          \\
				k5\_t8     & 8               & 5           
				& \ding{55}          \\
				k5\_t8\_dw & 8               & 5           
				& \checkmark          \\
				k7\_t4     & 4               & 7           
				& \ding{55}          \\
				k7\_t4\_dw & 4               & 7           
				& \checkmark          \\
				k7\_t8     & 8               & 7           
				& \ding{55}          \\
				k7\_t8\_dw & 8               & 7           
				& \checkmark          \\ \hline
			\end{tabular}      
		\end{threeparttable} 
	}
\end{table}

\subsection{MCAS-relaxed Supernet}
As discussed in Section~\ref{subsection:Refine_Fusion_MLP_Module}, our MCAS enable DNAS to optimize various configurations of the Wave-MLP block within each basic layer, while decoupling the optimization of channel numbers across different harmonization blocks.
Both optimization processes are realized through Harmony-relaxation technique, as explained in Section~\ref{subsubsec:relax_across} and Section~\ref{subsubsec:relax_within} below. The rationale for this design, discussed in Section \ref{subsubsection:computational_analysis}, is that the number of channels is the dominant factor affecting the FLOPs and parameter count of stacked Wave-MLP blocks. Consequently, the derived RFM module effectively addresses challenge (2) from Section \ref{subsection:vision_mlp}.

\subsubsection{Relaxation across Harmonization Blocks}
\label{subsubsec:relax_across}

Consider the block $H_i$, which receives feature maps from its $m$ preceding blocks, denoted as $H_{i-m}, H_{i-m+1}, \dots, H_{i-1}$. The CFM is also treated as $H_0$. Crucially, each harmonization block has a fixed channel dimension (see Table \ref{tab:ch_ph} for channel configurations of $H_1$, $H_2$, and $H_3$), where $H_0$ is fixed to 256 channels. The channel dimensions of input feature maps are aligned to uniform sizes via CH layers. The output of the CH layers in $H_i$ is computed using softmax-based relaxation:
\begin{equation}
    \text{output} = 
    \sum_{k=1}^m 
    \frac{{\rm exp} (\beta_{i-k,i})}
    {\sum_{k'=1}^{m'} {{\rm exp} (\beta_{i-k,k'})}}
    \cdot C_{ik}
    (H_{i-k}\text{'s output}).
    \nonumber
\end{equation}
Here, $\beta_{i-k,i}$ is a learnable architecture parameter for the path from $H_{i-k}$ to $H_i$, and $m'$ denotes the number of subsequent harmonization blocks connected to $H_{i-k}$.
This relaxation mechanism directly optimizes the channel configuration by selecting the optimal inter-block connections, where each connection path corresponds to a specific fixed channel dimension. After training, the final channel configuration is determined by selecting the path with the maximum $\beta_{i-k,i}$ value (i.e., $\arg\max_{k} \beta_{i-k,i}$) for each connection, which inherently resolves the effective channel numbers across the entire network.
For example, in the final architecture, $H_3$ will only accept inputs from $H_1$ or $H_2$ (whichever has larger $\beta$), automatically selecting between $256\to 384$ or $320\to 384$ channel transformations.

\subsubsection{Relaxation within Basic Layer}
\label{subsubsec:relax_within}

Let $\mathcal{O}^c$ denote the set of Wave-MLP blocks with different configurations (e.g., kernel size, MLP expansion ratio) but \textit{identical} channel dimensions (fixed by CH layers' connection decisions). A trainable architecture parameter ${\alpha}_o^l$ is assigned to each candidate $o \in \mathcal{O}^c$ in basic layer $l$. The output of basic layer $l$ is computed as:
\begin{equation}
    \text{output} = 
    \sum_{o\in \mathcal{O}^c}{
        \frac{{\rm exp}({\alpha}_o^l)}{
            \sum_{o'\in \mathcal{O}^c}{{\rm exp}({\alpha}_{o'}^l)}}
        \cdot o(\text{input})}.
    \nonumber
\end{equation}
Importantly, within each basic layer, the channel dimension is already fixed (as detailed above). 
This relaxation only selects the optimal operation configuration (i.e., kernel size, expansion ratio and whether to use depthwise convolution) for the predetermined channel size. After training, the final operation for each basic layer is chosen via $\arg\max_{o} \alpha_o^l$ to extract the highest-weighted candidate. For instance, in $P_{31}$ (384-channel PH layer), the system selects between "wavemlp\_k5\_t4\_dw" or "wavemlp\_k7\_t8" based on $\alpha$ weights, while the channel count (384) remains unchanged.

\subsubsection{Evaluation of Possible Architectures}
After relaxing the MCAS search space into a continuous supernet, Table~\ref{tab:archi_num} enumerates the total candidate architecture count. The complete search space comprises approximately $1.83\times10^{13}$ distinct configurations for the Siamese tracker neck (comprising CFM and RFM modules), which aligns with the typical magnitude range for specialized DNAS methodologies in computer vision. A detailed visualization of MCAS and the supernet structure is provided in Appendix~\ref{append:search_space_and_derived_model}.

\begin{table}[h]
	\caption{Number of All Possible Architectures}
	\label{tab:archi_num}
	\centering
	\resizebox{0.6\linewidth}{!} 
	{
		\begin{threeparttable} 
			\begin{tabular}{cc}
				\hline
				Routing Path &
				Number of Architectures \\ \hline
				$0\to 1$ & 
				$12\times 13\times 13\times 13 = 26,364$ \\
				$0\to 1\to 2$ & 
				$(12\times 13\times 13\times 13)^2 = 695,060,496$ \\
				$0\to 2$ & 
				$12\times 13\times 13\times 13 = 26,364$ \\
				$0\to 1\to 2\to 3$ & 
				$(12\times 13\times 13\times 13)^3 = 18,324,574,916,544$ \\
				$0\to 1\to 3$ & 
				$(12\times 13\times 13\times 13)^2 = 695,060,496$ \\
				$0\to 2\to 3$ & 
				$(12\times 13\times 13\times 13)^2 = 695,060,496$ \\
				$0\to 3$ & 
				$12\times 13\times 13\times 13 = 26,364$                    
				\\ \hline
			\end{tabular}
			\begin{tablenotes}  
				\footnotesize          
				\item[1] The notation $0\to 2\to 3$ indicates a path that starts from $H_0$ and passes through $H_2$ before directly producing the final output at $H_3$. 
				Other routing paths follow the same convention.
			\end{tablenotes}  
		\end{threeparttable} 
	}
\end{table}

\subsection{Optimization Strategy}
\label{subsection:tsdc-dnas}
\begin{algorithm}[h]
	\caption{Optimization Procedure for MCAS}
	\label{alg:optimization}
	\begin{algorithmic}[1]
		\Require Dataset $\mathcal{D}_{total}=\mathcal{D}_{train}\cup\mathcal{D}_{val}$, architecture parameters $\boldsymbol{\alpha}$ and $\boldsymbol{\beta}$, model weight parameters $\boldsymbol{\gamma}$
		\Ensure Optimized parameters $\boldsymbol{\alpha}$, $\boldsymbol{\beta}$ and $\boldsymbol{\gamma}$
		
		\Statex \textbf{Stage 1: Warm up with shallow training}
		\State Freeze $\boldsymbol{\alpha}, \boldsymbol{\beta}$ and feature extractor (backbone) weights
		\For{each mini-batch in $\mathcal{D}$}
		\State Update $\boldsymbol{\gamma}$ by minimizing ${\mathcal{L}}_{total}(\boldsymbol{\gamma}, \boldsymbol{\alpha}, \boldsymbol{\beta})$
		\EndFor
		
		\Statex \textbf{Stage 2: Warm up with deep training}
		\State Unfreeze feature extractor weights, keep $\boldsymbol{\alpha}$ and $\boldsymbol{\beta}$ frozen
		\For{each mini-batch in $\mathcal{D}$}
		\State Update $\boldsymbol{\gamma}$ by minimizing ${\mathcal{L}}_{total}(\boldsymbol{\gamma}, \boldsymbol{\alpha}, \boldsymbol{\beta})$
		\EndFor
		
		\Statex \textbf{Stage 3: Alternate Optimization of $\boldsymbol{\gamma}$ and $\boldsymbol{\alpha}, \boldsymbol{\beta}$}
		\For{each epoch}
		\For{each mini-batch in $\mathcal{D}_{train}$}
		\State Update $\boldsymbol{\gamma}$ by minimizing ${\mathcal{L}}_{train}(\boldsymbol{\gamma}, \boldsymbol{\alpha}, \boldsymbol{\beta})$
		\EndFor
		
		\For{each mini-batch in $\mathcal{D}_{val}$}
		\State Update $\boldsymbol{\alpha}$ and $\boldsymbol{\beta}$ by minimizing ${\mathcal{L}}_{\text{val}}(\boldsymbol{\gamma}, \boldsymbol{\alpha}, \boldsymbol{\beta})$ using SGD \cite{bottou2010large} and Adam \cite{kingma2014adam}, respectively.
		\EndFor
		\EndFor
	\end{algorithmic}
\end{algorithm}
To derive a lightweight correlation architecture from the MCAS-relaxed supernet through training, we propose a novel optimization strategy specifically designed for tracking tasks. As shown in pseudocode Algorithm~\ref{alg:optimization}, the optimization procedure is divided into three stages.
The first stage aims to mitigate the risk of the feature extractor's weight parameters being adversely affected by the undertrained supernet during the initial epochs. The second stage addresses the tendency of the supernet to favor shallow architectures due to the limited representational capacity of the Wave-MLP blocks in the early stages of training. The third stage implements bi-level optimization where weight updates and architecture parameter updates are alternated to converge on Pareto-optimal solutions balancing accuracy and efficiency.
\paragraph{Loss Function}
The loss function used during the optimization process is also tailored for tracking tasks. The MCAS-derived supernet outputs one response map, which is then processed by the tracking head, decoding the outcomes of both regression and classification. We employ the Intersection over Union (IOU) loss \cite{yu2016unitbox} for regression, denoted as ${\mathcal{L}}_{reg}$, and binary cross-entropy (BCE) loss for classification, denoted as ${\mathcal{L}}_{cls}$. The total loss function during the search stage is formulated as:
\begin{equation}
    {\mathcal{L}}(\boldsymbol{\gamma},\boldsymbol{\alpha},\boldsymbol{\beta})=
    \eta{\mathcal{L}}_{sea} + \lambda{\mathcal{L}}_{reg} + \mu{\mathcal{L}}_{cls},
\end{equation}
where ${\mathcal{L}}_{sea}$ represents the inverse chained cost (defined in Appendix~\ref{append:inverse_chained_cost_estimation}) \cite{fang2020densely} of supernet's latency or FLOPs. The hyperparameters $\eta$, $\lambda$, and $\mu$ control the relative contribution of each term, respectively.

\paragraph{Optimizer Design}
We employ distinct optimizers for the optimization of channel-level ($\boldsymbol{\beta}$) and operation-level ($\boldsymbol{\alpha}$) parameters. Adam \cite{kingma2014adam} is used for $\boldsymbol{\beta}$ to stabilize sparse connection paths with high variance, while SGD \cite{bottou2010large} with momentum optimizes $\boldsymbol{\alpha}$ to enable robust exploration of operation candidates within fixed channel configurations. This decoupled optimization strategy efficiently navigates the hierarchical search space while maintaining computational tractability.

\section{Experiments}
\label{section:experiments}

\subsection{Configurations of Implementation}
The proposed SEAT tracker was searched and retrained on a computer with four NVIDIA 3090 GPUs. In the case of SEAT\_AL, the sizes of the template $\boldsymbol{t}$ and search region $\boldsymbol{s}$ were set to $127 \times 127$ and $255 \times 255$, respectively. AlexNet served as the feature extractor for our tracker, operating with a stride of 8. In the search stage, the MCAS-relaxed supernet is trained for 48 epochs according to the strategy outlined in Sec~\ref{subsection:tsdc-dnas}.
The weight parameters $\boldsymbol{\gamma}$ are optimized on the training dataset using an SGD \cite{bottou2010large} optimizer, with the learning rate exponentially decaying from $5\times 10^{-3}$ to $10^{-5}$ and employing a warm-up trick in the first 5 epochs. 
The architecture parameters $\boldsymbol{\alpha}$ and $\boldsymbol{\beta}$ were optimized on the validation dataset using both SGD and Adam \cite{kingma2014adam} optimizers. The Adam optimizer was initialized with a learning rate of $10^{-4}$ and a weight decay of $10^{-3}$.
The hyper-parameters $\eta,\lambda,\mu$ are set to 1. During the retrain progress, the model derived from the search stage is trained for 48 epochs in the same manner as AutoMatch~\cite{zhang2021learn}. The same weight optimizer and learning rate are utilized as in the search stage without inheriting any parameters from the search stage.

We utilize COCO \cite{lin2014microsoft}, YouTube-BB \cite{real2017youtube}, GOT-10k \cite{huang2019got}, as well as ImageNet DET and VID \cite{russakovsky2015imagenet} as the training dataset for experiments on four generic tracking benchmarks and three aerial tracking benchmarks.
To enable comparisons with both generic trackers on resource-constrained GPUs and UAV trackers on resource-limited NPUs, we derive two configurations of our model: SEAT\_LT and SEAT\_AL, respectively (Figure~\ref{fig_search} in Appendix~\ref{append:search_space_and_derived_model} visualizes these configurations). SEAT\_LT is further used in ablation studies to demonstrate the effectiveness of the proposed components in our correlation architecture, namely the CFM Module and the RFM Module.

\subsection{Evaluation Datasets and Metrics}
We conduct evaluations of SEAT\_LT on generic tracking benchmark datasets, namely GOT10K \cite{huang2019got}, OTB2015 \cite{7001050}, VOT2019 \cite{kristan2019seventh}, and NFS30 \cite{kiani2017need}. SEAT\_AL undergoes evaluation on a diverse set of aerial tracking benchmark datasets, including UAV123 \cite{mueller2016benchmark} (which includes the UAV10FPS and UAV20L), UAVDT \cite{du2018unmanned}, and VISDRONE \cite{fan2020visdrone}.

\subsubsection{Generic Benchmarks.} \textbf{GOT10K}: A substantial dataset featuring over 10,000 videos capturing moving objects in real-world scenarios. Notably, it poses a challenge with zero-class-overlap between the training and testing subsets. \textbf{VOT2019}: Featuring 60 challenging sequences, VOT2019 measures tracking accuracy and robustness concurrently using the expected average overlap (EAO) metric. \textbf{NFS30}: A long-range benchmark consisting of 100 videos (380,000 frames) captured with higher frame rate (240FPS) cameras, providing real-world scenarios. \textbf{OTB2015}: Comprising 100 video sequences, OTB2015 introduces challenging attributes that reflect various visual effects leading to tracking difficulties. The average sequence length is approximately 500 frames.

\subsubsection{Aerial Benchmarks.} \textbf{UAV10FPS}: This dataset involves video sequences derived from the original 30fps version by selecting every 10 frames. Consequently, UAV10FPS introduces more severe motion challenges compared to UAV123~\cite{mueller2016benchmark}. \textbf{UAV20L}: Featuring 20 long-term video sequences created by merging short subsequences from UAV123, UAV20L presents progressively larger and irregular changes in object location between frames, posing a heightened challenge for object tracking. \textbf{UAVDT}: Specifically designed to address complicated circumstances such as weather variations, altitude changes, diverse camera views, vehicle types, and occlusion, UAVDT provides a comprehensive evaluation of UAV tracking performance. \textbf{VISDRONE}: Originating from the Vision Meets Drone Single-Object Tracking challenge, VISDRONE utilizes various drone platforms across different weather and lighting conditions, providing a challenging and realistic evaluation environment.

\subsubsection{Evaluation Metrics} In our quantitative evaluation, we adhere to the specific protocols established by each tracking dataset to ensure a rigorous and consistent assessment of the tracker's performance. For all benchmarks, excluding VOT2019, we employ success plots as our primary evaluation metric. The success plot illustrates the percentage of successful frames across varying overlap thresholds, ranging from 0 to 1. A frame is deemed successful if the overlap between the predicted and ground truth bounding boxes surpasses a predefined threshold. The area under the success plot curve (AUC) serves as a quantitative measure, capturing the tracker's performance across different levels of overlap. In the case of the VOT2019 dataset, we compute the expected average overlap (EAO), a metric that provides a balanced assessment by considering both tracking accuracy and robustness.

\subsection{Trackers for Lightweight GPU-based Applications}
In Table~\ref{tab_speed_generic_performance}, which presents a comprehensive comparison of tracker performances, SEAT\_LT emerges as an exemplary lightweight tracker. It not only demonstrates superior accuracy compared to other lightweight generic trackers such as LightTrack \cite{yan2021lighttrack}, FEAR \cite{borsuk2022fear}, and E.T.Track \cite{blatter2023efficient} across various benchmarks, securing top-tier rankings as indicated by the results highlighted in \textcolor{red}{red} and \textcolor{blue}{blue}, but also maintains a fast processing speed that ensures it remains within the lightweight category on resource-constrained GPU hardware.

\begin{table*}[h]
	\caption{Comparison of Trackers for Lightweight GPU Use on OTB2015, GOT10KTEST and VOT2019}
	\label{tab_speed_generic_performance}
	\centering
	\resizebox{1.0\linewidth}{!} 
	{
		\begin{tabular}{c|ccc|ccccccc}
			\hline
			&                                                 &                              &                               & \multicolumn{2}{c}{VOT2019}                                                                                           &                                                                      &                                                                         &                                                                    & \multirow{2}{*}{\begin{tabular}[c]{@{}c@{}}FPS\\ ↑\end{tabular}} & \multirow{2}{*}{\begin{tabular}[c]{@{}c@{}}Latency\\ ↓ (ms)\end{tabular}} \\ 
			\cline{5-6}
			\multirow{-2}{*}{\begin{tabular}[c]{@{}c@{}}Orientation\end{tabular}} 
			& \multirow{-2}{*}{Trackers}                      
			& \multirow{-2}{*}{Year}       
			& \multirow{-2}{*}{Type}        
			& EAO↑ & Robustness↓
			& \multirow{-2}{*}{\begin{tabular}[c]{@{}c@{}}OTB\\ 2015\end{tabular}} 
			& \multirow{-2}{*}{\begin{tabular}[c]{@{}c@{}}GOT10K\\ TEST\end{tabular}} 
			& \multirow{-2}{*}{\begin{tabular}[c]{@{}c@{}}NFS\\ 30\end{tabular}} 
			&  & \\ \hline
			
			& {\color[HTML]{656565} ECO}                    
			& {\color[HTML]{656565} 2017}  
			& {\color[HTML]{656565} -}      
			& {\color[HTML]{656565} -}      
			& {\color[HTML]{656565} -}      
			& {\color[HTML]{656565} 64.3}    
			& {\color[HTML]{656565} -}        
			& {\color[HTML]{656565} 46.6}    
			& {\color[HTML]{656565} -}      
			& {\color[HTML]{656565} -}      \\
			
			& \cellcolor[HTML]{E6E6E6}LightTrack(M)           
			& \cellcolor[HTML]{E6E6E6}2021 
			& \cellcolor[HTML]{E6E6E6}cnn   
			& \cellcolor[HTML]{E6E6E6}{\color[HTML]{3531FF} {\ul 33.3}} 
			& \cellcolor[HTML]{E6E6E6}{\color[HTML]{3531FF} {\ul 32.1}} 
			& \cellcolor[HTML]{E6E6E6}65.8     
			& \cellcolor[HTML]{E6E6E6}61.1      
			& \cellcolor[HTML]{E6E6E6}56.3     
			& \cellcolor[HTML]{E6E6E6}50.20 
			& \cellcolor[HTML]{E6E6E6}19.9 \\
			
			& \cellcolor[HTML]{E6E6E6}FEAR(XS)                
			& \cellcolor[HTML]{E6E6E6}2022 
			& \cellcolor[HTML]{E6E6E6}cnn   
			& \cellcolor[HTML]{E6E6E6}25.2  
			& \cellcolor[HTML]{E6E6E6}56.2  
			& \cellcolor[HTML]{E6E6E6}66.7     
			& \cellcolor[HTML]{E6E6E6}{\color[HTML]{3531FF} {\ul 61.9}} 
			& \cellcolor[HTML]{E6E6E6}58.3     
			& \cellcolor[HTML]{E6E6E6}{\color[HTML]{FE0000} {\ul 123.33}} 
			& \cellcolor[HTML]{E6E6E6}{\color[HTML]{FE0000} {\ul 8.1}} \\
			
			& \cellcolor[HTML]{E6E6E6}E.T.Track               
			& \cellcolor[HTML]{E6E6E6}2023 
			& \cellcolor[HTML]{E6E6E6}trans 
			& \cellcolor[HTML]{E6E6E6}24.1  
			& \cellcolor[HTML]{E6E6E6}57.2  
			& \cellcolor[HTML]{E6E6E6}{\color[HTML]{3531FF} {\ul 67.8}} 
			& \cellcolor[HTML]{E6E6E6}58.3     
			& \cellcolor[HTML]{E6E6E6}{\color[HTML]{3531FF} {\ul 59.0}} 
			& \cellcolor[HTML]{E6E6E6}48.15 
			& \cellcolor[HTML]{E6E6E6}20.8 \\
			
			\multirow{-5}{*}{\begin{tabular}[c]{@{}c@{}}GPU-constrained\\ platforms\end{tabular}} 
			& \cellcolor[HTML]{E6E6E6}\textbf{SEAT\_LT(ours)} 
			& \cellcolor[HTML]{E6E6E6}2023 
			& \cellcolor[HTML]{E6E6E6}mlp   
			& \cellcolor[HTML]{E6E6E6}{\color[HTML]{FE0000} {\ul 34.0}} 
			& \cellcolor[HTML]{E6E6E6}{\color[HTML]{FE0000} {\ul 31.6}} 
			& \cellcolor[HTML]{E6E6E6}{\color[HTML]{FE0000} {\ul 69.7}} 
			& \cellcolor[HTML]{E6E6E6}{\color[HTML]{FE0000} {\ul 63.6}} 
			& \cellcolor[HTML]{E6E6E6}{\color[HTML]{FE0000} {\ul 61.1}} 
			& \cellcolor[HTML]{E6E6E6}{\color[HTML]{3531FF} {\ul 80.54}} 
			& \cellcolor[HTML]{E6E6E6}{\color[HTML]{3531FF} {\ul 12.4}} \\ \hline
		\end{tabular}
	}
\end{table*}

Table~\ref{tab_speed_generic_performance} demonstrates that SEAT\_LT achieves an impressive balance between accuracy, model efficiency, and GPU performance. On the OTB2015 benchmark, it achieves a success rate of 69.7\%, surpassing its closest competitor, E.T.Track, by 1.9\%. On the VOT2019 dataset, it continues to lead with the highest Expected Average Overlap (EAO) score of 34.0\%, slightly ahead of LightTrack's 33.3\%.
In terms of computational efficiency, SEAT\_LT operates with just 12.28 million parameters and 6.01 billion FLOPs, significantly lighter than more resource-intensive trackers, such as AutoMatch \cite{zhang2021learn} and TransT \cite{chen2021transformer}, which typically require around 20 million parameters and 20 billion FLOPs. Moreover, SEAT\_LT achieves an impressive frame rate of 80.54 FPS, making it highly suitable for real-time tracking applications. SEAT\_LT proves to be an ideal solution for resource-constrained GPU environments.

\begin{figure}[h]
	\centering
	\includegraphics[width=8.8cm]{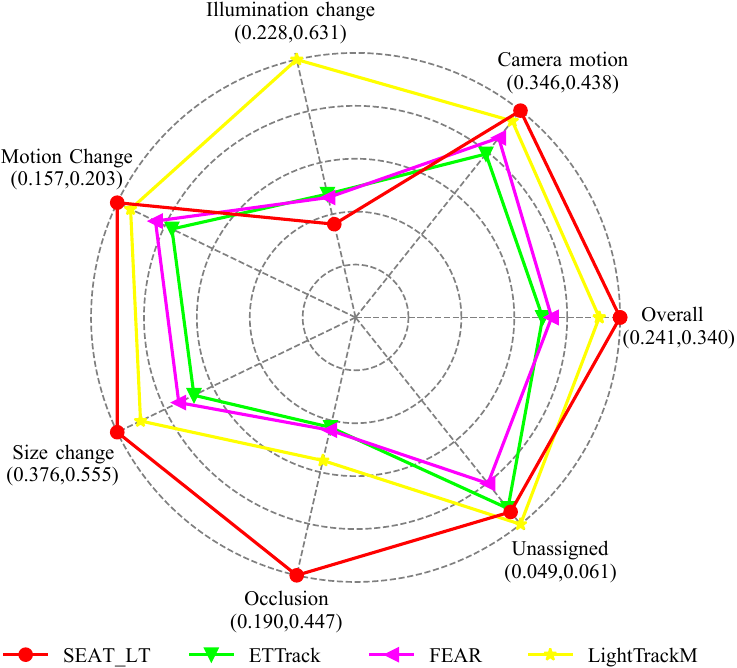}
	\caption{
		Comparative analysis of tracker attributes on VOT2019. The radar chart annotates the minimum and maximum EAO scores per attribute. SEAT\_LT attains consistently high EAO scores across attributes—except for illumination change—and surpasses other lightweight GPU-oriented trackers, with particularly strong gains under occlusion.
	}
	\label{fig_attr_lt}
\end{figure}

In addition, the tracking efficacy of SEAT\_LT has been rigorously evaluated on the VOT2019 dataset, focusing on five critical attributes: camera motion, illumination changes, motion changes, size changes, and occlusion. As shown in Figure~\ref{fig_attr_lt}, SEAT\_LT consistently outperforms SOTA Siamese-based trackers optimized for lightweight GPU use, especially when compared to LightTrack.
The vertex annotations highlight SEAT\_LT’s consistent superiority across most attributes, with the exception of illumination changes. Its MLP-based fusion module leverages global dependencies captured by fully connected layers, enabling more effective target–background separation and improving tracking accuracy under various appearance variations. 
Performance degrades under severe illumination changes, particularly when the target blends into dark backgrounds. This is primarily due to the amplitude branch of the PATM module, which encodes intensity information and is therefore more sensitive to lighting variations. In practice, however, modern imaging sensors often apply illumination pre-processing, resulting in illumination-stable inputs that help mitigate this issue. Notably, SEAT\_LT performs robustly on complex sequences with challenging occlusions and outperforms other SOTA lightweight trackers in this attribute.

\subsection{Trackers for On-board NPU Applications}
As NPUs become increasingly essential for aerial tracking in on-board computing systems, we introduce SEAT\_AL as a novel solution tailored for UAV trackers operating in NPU-equipped edge environments. This approach leverages the same backbone architecture (AlexNet), which used by on-board UAV trackers such as SiamAPN \cite{fu2021siamese}, enabling direct performance comparisons while showcasing the distinct advantages of our method in edge computing scenarios.

\begin{figure*}[h]
	\centering
	\includegraphics[width=\linewidth]{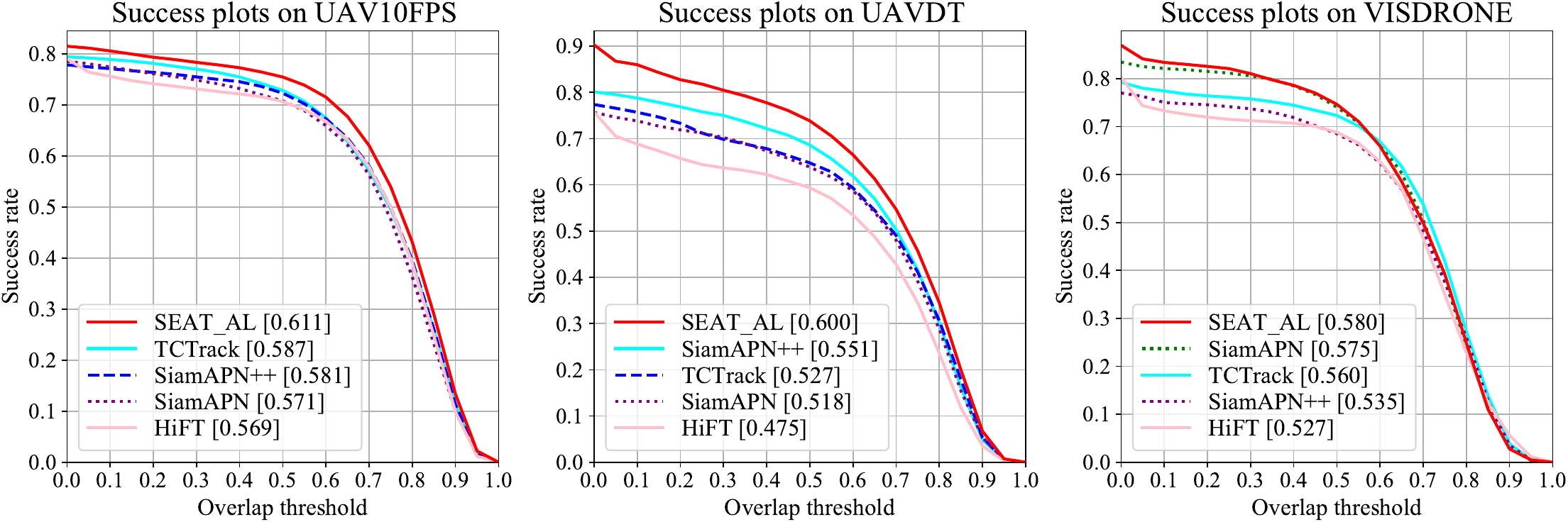}
	\caption{The comprehensive assessment of trackers tailored for on-board deployment encompasses three renowned aerial tracking benchmarks.}
	\label{fig_al_succ}
\end{figure*}

\subsubsection{Overall Performance}
The success plots depicted in Figure~\ref{fig_al_succ} showcase the performance of various trackers across three aerial benchmarks. The accompanying legend sorts trackers by their respective Area Under the Curve (AUC) metric. Our SEAT\_AL tracker demonstrates superior performance when compared to traditional tracking algorithms: SEAT\_AL achieves the highest success scores, leading the benchmarks with results of 0.611, 0.600 and 0.580 respectively. 

\begin{table*}[h]
	\caption{COMPARISON of MODEL DEPLOYMENT on HORIZON X3 CHIP among DIFFERENT ON-BOARD TRACKERS}
	\label{tab_deploy}
	\centering
	\resizebox{1.0\linewidth}{!} 
	{
		\begin{threeparttable} 
			\begin{tabular}{lccccccc}
				\toprule
				& MACs & Params & FPS & Average Latency & Subgraphs & DDR Latency & UAV20L Accuracy \\
				\midrule
				SEAT\_AL(\textbf{ours})  
				&8.13G    &11.47M  &20.41  &195.47ms  &13 &892.43ms & 0.593\\
				\midrule
				SiamAPN++ \cite{cao2021siamapn++}
				&10.25G\textcolor{red}{$\uparrow$}    &15.15M\textcolor{red}{$\uparrow$}  &23.57\textcolor{red}{$\uparrow$}     &168.64ms\textcolor{red}{$\downarrow$}      &8\textcolor{red}{$\downarrow$}       &624.77ms\textcolor{red}{$\downarrow$}
				& 0.533\textcolor{red}{$\downarrow$}   \\
				HiFT \cite{cao2021hift}
				&4.83G\textcolor{red}{$\downarrow$}    &9.98M\textcolor{red}{$\downarrow$}  &4.30\textcolor{red}{$\downarrow$}     &921.41ms\textcolor{red}{$\uparrow$}      &28\textcolor{red}{$\uparrow$}       &681.40ms\textcolor{red}{$\downarrow$}
				& 0.566\textcolor{red}{$\downarrow$}   \\
				TCTrack \cite{cao2022tctrack}
				&3.37G\textcolor{red}{$\downarrow$}    &7.09M\textcolor{red}{$\downarrow$}  &0.71\textcolor{red}{$\downarrow$}     &5608.77ms\textcolor{red}{$\uparrow$}      &52\textcolor{red}{$\uparrow$}       &3461.16ms\textcolor{red}{$\uparrow$}
				& 0.516\textcolor{red}{$\downarrow$}   \\
				\bottomrule
			\end{tabular}
			\begin{tablenotes}  
				\footnotesize          
				\item[1] MACs (Multiply-Accumulate Operations) are a key metric for evaluating the computational complexity of deep learning models.
			\end{tablenotes}       
		\end{threeparttable} 
	}
\end{table*}

\subsubsection{Real-world Tests}
The efficacy of the discussed on-board trackers, primarily designed for UAV applications, was demonstrated through their deployment on the Horizon X3 chip, a low-power embedded platform with an NPU offering 5 TOPS for inference, along with 4 ARM A53 cores. This setup enabled a performance comparison between SEAT\_AL and other trackers.
As shown in Table \ref{tab_deploy}, SEAT\_AL strikes an impressive balance between speed and accuracy, achieving 20.41 FPS with an average latency of 195.47ms. It also boasts a relatively low number of subgraphs (13) and moderate DDR latency (892.43ms), while maintaining a high accuracy score (0.593) on the UAV20L benchmark. In contrast, trackers like HiFT and TCTrack, which rely on ViT-based correlation operations, exhibit significantly higher latencies and lower FPS due to the computationally heavy architecture, making them less efficient for real-world deployment on the Horizon X3.

\begin{figure*}[h]
	\centering
	\includegraphics[width=\linewidth]{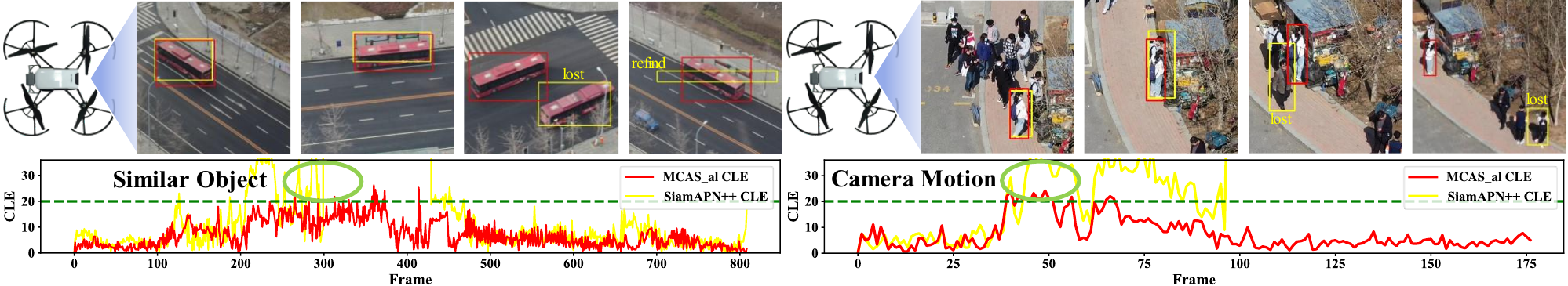}
	\caption{In real-world scenarios involving bus (left) and pedestrian (right), SEAT\_AL successfully addresses challenges related to similar objects and camera motion, maintaining robust tracking without any target loss. In contrast, SiamAPN++ \cite{cao2021siamapn++} struggles, losing the target in both cases.}
	\label{real-world}
	\vspace{-0.4cm}
\end{figure*}

Comprehensive real-world tests, depicted in Figure~\ref{real-world}, were conducted in two aerial tracking scenarios: bus (long-term tracking) and pedestrian (short-term tracking). Precision was evaluated using the Center Location Error (CLE), which measures the Euclidean distance between the ground truth box's center and the predicted box's center. The CLE curves were plotted, and points where the CLE exceeded the 20-pixel threshold were analyzed.
SEAT\_AL effectively handled challenges such as similar objects and camera motion, maintaining continuous and robust tracking throughout both scenarios. In contrast, SiamAPN++ struggled with target loss. Specifically, it lost and relocated the bus due to interference from a similar object and failed entirely to track the pedestrian due to camera motion.
The CLE curves consistently demonstrated SEAT\_AL's resilience in maintaining accurate tracking, while SiamAPN++ frequently experienced lapses in performance despite its higher FPS.
These real-world tests substantiate SEAT\_AL’s robustness and efficiency for UAV-specific tracking, confirming its suitability for on-board deployment. Furthermore, its stable accuracy–efficiency trade-off indicates strong potential for extension to indoor drone tracking scenarios \cite{10172187}.

\subsection{Ablation Studies}

We conduct comprehensive ablation studies to dissect the impact of key components in our proposed MLP-based Correlation Architecture Space. Our ablation targets include the Coarse Fusion MLP (CFM) Module and the Wave-MLP blocks within the Refine Fusion MLP (RFM) Module. 


\begin{table*}[h]
	\caption{ABLATION ANALYSIS of THE PROPOSED MODULES on GOT10K, OTB2015 and VOT2019}
	\label{tab_ablation}
	\centering
	\resizebox{0.95\linewidth}{!} 
	{
		\begin{tabular}{ccccc|ccccc}
			\hline
			&                                                                         &                                                                          &                                                                          &                                                                          & \multicolumn{5}{c}{SEAT\_LT}                                                                                                                                                                                                                                                                                                                     \\ \cline{6-10} 
			&                                                                         &                                                                          &                                                                          &                                                                          &                                                                           &                                                                             &                                                                             &                               &                                                                          \\
			\multirow{-3}{*}{\begin{tabular}[c]{@{}c@{}}Ablation\\ Study\end{tabular}} & \multirow{-3}{*}{\begin{tabular}[c]{@{}c@{}}Pre-\\ Fusion\end{tabular}} & \multirow{-3}{*}{\begin{tabular}[c]{@{}c@{}}Fusion\\ -Type\end{tabular}} & \multirow{-3}{*}{\begin{tabular}[c]{@{}c@{}}Post-\\ Fusion\end{tabular}} & \multirow{-3}{*}{\begin{tabular}[c]{@{}c@{}}Fusion\\ -Type\end{tabular}} & \multirow{-2}{*}{\begin{tabular}[c]{@{}c@{}}GOT10K\\ (AO)\%\end{tabular}} & \multirow{-2}{*}{\begin{tabular}[c]{@{}c@{}}OTB2015\\ (AUC)\%\end{tabular}} & \multirow{-2}{*}{\begin{tabular}[c]{@{}c@{}}VOT2019\\ (EAO)\%\end{tabular}} & \multirow{-2}{*}{FPS}         & \multirow{-2}{*}{\begin{tabular}[c]{@{}c@{}}Latency\\ (ms)\end{tabular}} \\ \hline
			Ab1                                                                        & cnn                                                                     & patch                                                                    & mlp                                                                      & pixel                                                                    & 52.2                                                                      & 62.8                                                                        & 22.3                                                                        & 69.63                         & 14.4                                                                     \\
			Ab2                                                                        & transformer                                                             & pixel                                                                    & mlp                                                                      & pixel                                                                    & 62.6                                                                      & 65.7                                                                        & 30.8                                                                        & 61.23                         & 16.3                                                                     \\ \hline
			Ab0                                                                        & mlp                                                                     & pixel                                                                    & -                                                                        & -                                                                        & 63.5                                                                      & 68.1                                                                        & {\color[HTML]{3531FF} 33.7}                                                 & 89.39                         & 11.2                                                                     \\
			Ab3                                                                        & mlp                                                                     & pixel                                                                    & cnn                                                                      & patch                                                                    & 63.1                                                                      & {\color[HTML]{3531FF} 69.6}                                                 & 31.5                                                                        & 83.82                         & 11.9                                                                     \\
			Ab4                                                                        & mlp                                                                     & pixel                                                                    & transformer                                                              & pixel                                                                    & {\color[HTML]{FE0000} {\ul 63.8}}                                         & {\color[HTML]{3531FF} 69.6}                                                 & 33.6                                                                        & 77.99                         & 12.8                                                                     \\ \hline
			ours                                                                       & mlp                                                                     & pixel                                                                    & mlp                                                                      & pixel                                                                    & {\color[HTML]{3531FF} 63.6}                                               & {\color[HTML]{FE0000} {\ul 69.7}}                                           & {\color[HTML]{FE0000} {\ul 34.0}}                                           & \cellcolor[HTML]{EFEFEF}80.54 & \cellcolor[HTML]{EFEFEF}12.4                                             \\ \hline
		\end{tabular}
	}
\end{table*}

\begin{figure*}[h]
	\centering
	\includegraphics[width=\linewidth]{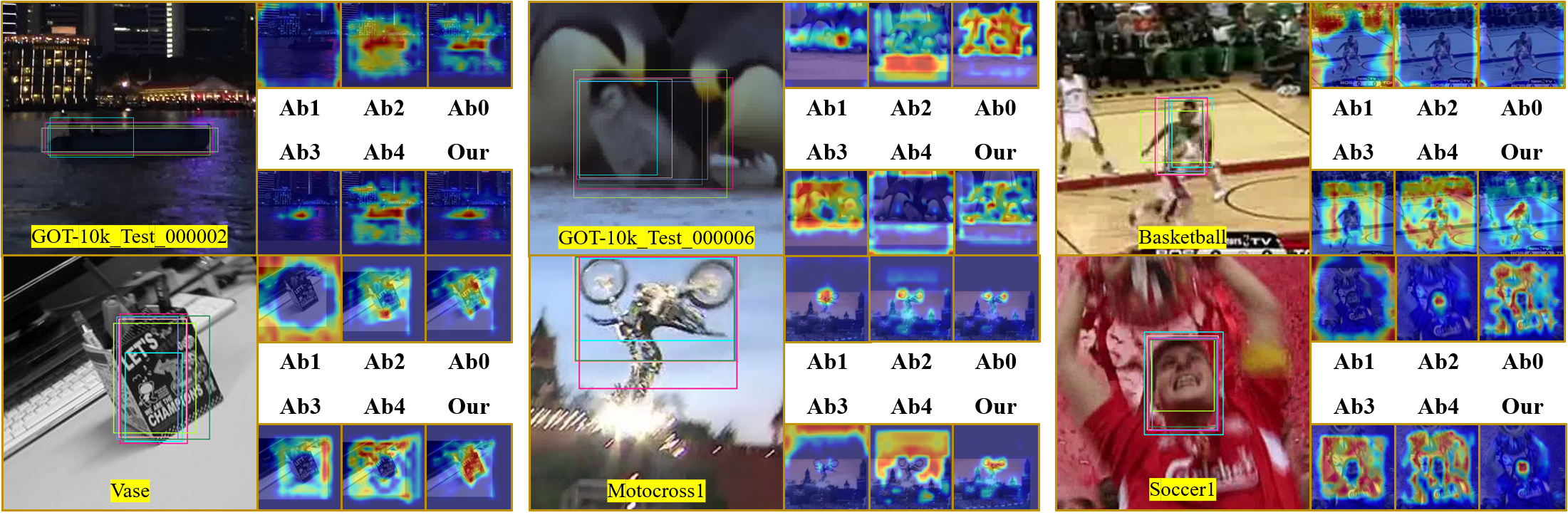}
	\caption{Visualization of ablation study. We adopt EigenCAM~\cite{muhammad2020eigen} to visualize of feature maps outputted by collector layer in ablation experiments of SEAT\_LT. Original images are chosen from GOT10KTEST, OTB2015 and VOT2019. Our tracker yields feature maps whose attention gathers at target region better.}
	\label{fig_abla_vis}
\end{figure*}

Table \ref{tab_ablation} provides a detailed breakdown of results for each component and their combinations across five ablation experiments (Ab0-4). We explore different pre-fusion operations, including CFM, Cross-Attention, and Depthwise-Convolution, representing mlp-based fusion, transformer-based fusion, and cnn-based fusion, respectively. The "Fusion Type" denotes whether fusion occurs at the pixel or patch level. Additionally, Figure~\ref{fig_abla_vis} presents visualizations of the ablation studies.

From the comparison between Ab1, Ab2, and ours in Table \ref{tab_ablation} and Figure~\ref{fig_abla_vis}, the effectiveness of CFM is evident, showcasing higher AUC and EAO across three benchmark datasets. While adopting Depthwise Convolution as a pre-fusion operation yields higher FPS, CFM exhibits superior performance with a modest speed cost.
To further illustrate the balance between speed and accuracy achieved by our RFM, we replace Wave-MLP blocks with ViT blocks and ResNet blocks, varying architectural types and fusion strategies. Ab3, Ab4, and ours in Table \ref{tab_ablation} and Figure~\ref{fig_abla_vis} demonstrate the higher FPS of Wave-MLP-based RFM compared to ViT-block-based post-fusion, along with higher accuracy than ResNet-block-based post-fusion and more focused concentration on target region. 

Notably, Ab0 does not involve any post-fusion operation, highlighting two key points: 1) The efficacy of our CFM design is evident. Through the decomposition of target and background features and their integration with search features in an attention-like manner, Ab0 achieves impressive performance at high speed, despite the feature map's visualization showing a lack of concentration.
2) The efficacy of transitioning from pre-fusion to post-fusion (coarse-to-fine) is demonstrated by the superior performance of our SEAT compared to Ab0. This indicates that adopting the proposed coarse-to-refine framework (CFM+RFM in MCAS) for fusing template and search feature maps enables the tracker to focus more effectively on the target region. Consequently, this approach mitigates limitations associated with small components of the target or distractions from background noise.

\section{Conclusion}
\label{section:conclusion}
We propose SEAT, an MLP-based tracker that addresses the long-standing efficiency-accuracy trade-off in Siamese visual object tracking, with a particular focus on resource-constrained settings. 
By fundamentally redesigning feature fusion in the Siamese neck, we introduce the CFM and RFM modules and realize a simple yet effective coarse-to-fine fusion framework. This design enables pixel-level feature fusion via MLPs, yielding improved accuracy.
To avoid the quadratic complexity with respect to channel width that arises from stacking Wave-MLP blocks in RFM, we construct the MCAS search space by jointly grouping CFM and RFM into a unified supernet under a Harmony-Relaxation strategy. Within this DNAS framework, channel width is decoupled from other structural choices: channel counts are optimized across Harmonization blocks using Adam optimizer, while intra-block architectural parameters are optimized with SGD. Our search identifies effective Wave-MLP configurations for RFM, improving the efficiency of the derived models. 
Consequently, we develop two lightweight trackers: SEAT\_LT, optimized for resource-constrained GPUs, and SEAT\_AL, tailored to low-power NPUs. Both achieve SOTA accuracy with superior inference speed.
Looking ahead, future work will focus on evolving SEAT into a one-stage architecture, integrating backbone feature extraction and neck feature fusion into a unified process (for searching). This evolution is expected to enhance accuracy in resource-abundant GPU environments while further simplifying model design for improved scalability and efficiency.

\section*{Acknowledgments}
We are sincerely grateful to Lu Huang, a Ph.D. candidate at Peking University, for her work in creating the figures and other visual materials for this paper. We also thank Prof. Ding Yuan (School of Astronautics, Beihang University) and Prof. Muhammad Haris Khan (MBZUAI) for their early guidance on the writing and presentation of the manuscript. In addition, we appreciate Zhaoxi Li and Guosheng Xu from Horizon Robotics for their technical assistance and valuable guidance on deploying the tracking algorithm on the Horizon X3 chip.

\bibliographystyle{unsrt}  
\bibliography{references}  

\appendix

\section{More Details of Methodology}
\label{append:more_details_of_methodology}

\subsection{Computational Analysis of Wave-MLP}
\label{append:computational_analysis_of_wavemlp}

The Wave-MLP \cite{tang2022image} blocks incorporate Phase-Aware Token Mixing (PATM) to fuse all feature tokens and enhance their representations. Upon obtaining the amplitude representation $\boldsymbol{h}$ and phase representation $\boldsymbol{\theta}$ for all tokens, given by (both $\boldsymbol{h}_\phi$ and $\boldsymbol{\theta}_\phi$ for $\phi=1,2,\cdots,H_sW_s$ are all row vectors with $C'$ length):
\begin{equation}
	\boldsymbol{h} =
	\begin{bmatrix}
		\boldsymbol{h}_1^\top & 
        \boldsymbol{h}_2^\top & 
        \cdots & 
        \boldsymbol{h}_{H_sW_s}^\top
	\end{bmatrix}^\top
    \in \mathbb{R}^{H_sW_s\times C'}, 
    \qquad
	\boldsymbol{\theta} =
	\begin{bmatrix}
		\boldsymbol{\theta}_1^\top & 
        \boldsymbol{\theta}_2^\top & 
        \cdots & 
        \boldsymbol{\theta}_{H_sW_s}^\top
	\end{bmatrix}^\top
    \in \mathbb{R}^{H_sW_s\times C'},
	\nonumber
\end{equation}
the enhanced consequence $\boldsymbol{o}$ can be computed using the Euler formula with trainable parameters:
\begin{align}
	&S(\boldsymbol{o})_{C'H_sW_s\times1} =
	\begin{bmatrix}
		\boldsymbol{o}_1^\top & 
		\boldsymbol{o}_2^\top & 
		\cdots & 
		\boldsymbol{o}_{H_sW_s}^\top
	\end{bmatrix}^\top
	\nonumber \\
	=&
	\begin{bmatrix}
		\boldsymbol{W}_{1,1}^{cos} & 
		\cdots & 
		\boldsymbol{W}_{1,H_sW_s}^{cos} &
		\boldsymbol{W}_{1,1}^{sin} & 
		\cdots & 
		\boldsymbol{W}_{1,H_sW_s}^{sin} \\
		\vdots                    & 
		\ddots & 
		\vdots                    &
		\vdots                    & 
		\ddots & 
		\vdots                    \\
		\boldsymbol{W}_{H_sW_s,1}^{cos} & 
		\cdots & 
		\boldsymbol{W}_{H_sW_s,H_sW_s}^{cos} &
		\boldsymbol{W}_{H_sW_s,1}^{sin} & 
		\cdots & 
		\boldsymbol{W}_{H_sW_s,H_sW_s}^{sin}
	\end{bmatrix}
	\begin{bmatrix}
		\boldsymbol{h}_1^\top 
        \odot 
        \left(\cos\boldsymbol{\theta}_1\right)^\top \\
		\vdots                                                   \\
		\boldsymbol{h}_{H_sW_s}^\top  
        \odot 
        \left(\cos\boldsymbol{\theta}_{H_sW_s}\right)^\top \\
		\boldsymbol{h}_1^\top  
        \odot 
        \left(\sin\boldsymbol{\theta}_1\right)^\top \\
		\vdots                                                   \\
		\boldsymbol{h}_{H_sW_s}^\top  
        \odot 
        \left(\sin\boldsymbol{\theta}_{H_sW_s}\right)^\top
	\end{bmatrix} 
	\nonumber \\ 
	=&
	\begin{bmatrix}
		\boldsymbol{W}_{1,1}^{cos} & 
		\cdots & 
		\boldsymbol{W}_{1,H_sW_s}^{cos} &
		\boldsymbol{W}_{1,1}^{sin} & 
		\cdots & 
		\boldsymbol{W}_{1,H_sW_s}^{sin} \\
		\vdots                    & 
		\ddots & 
		\vdots                    &
		\vdots                    & 
		\ddots & 
		\vdots                    \\
		\boldsymbol{W}_{H_sW_s,1}^{cos} & 
		\cdots & 
		\boldsymbol{W}_{H_sW_s,H_sW_s}^{cos} &
		\boldsymbol{W}_{H_sW_s,1}^{sin} & 
		\cdots & 
		\boldsymbol{W}_{H_sW_s,H_sW_s}^{sin}
	\end{bmatrix}
    \left\{
	\begin{bmatrix}
		\boldsymbol{h}_1^\top \\
		\vdots             \\
		\boldsymbol{h}_{H_sW_s}^\top \\
		\boldsymbol{h}_1^\top \\
		\vdots             \\
		\boldsymbol{h}_{H_sW_s}^\top
	\end{bmatrix} 
	\odot
	\begin{bmatrix}
		\left(\cos\boldsymbol{\theta}_1\right)^\top \\
		\vdots                           \\
		\left(\cos\boldsymbol{\theta}_{H_sW_s}\right)^\top \\
		\left(\sin\boldsymbol{\theta}_1\right)^\top \\
		\vdots                           \\
		\left(\sin\boldsymbol{\theta}_{H_sW_s}\right)^\top
	\end{bmatrix} 
    \right\}
	\nonumber \\
	=&
	\begin{bmatrix}
		\boldsymbol{W}^{cos} & \boldsymbol{W}^{sin}
	\end{bmatrix}
	\begin{bmatrix}
		S(\boldsymbol{h}) \odot {\rm cos}S(\boldsymbol{\theta})  \\
		S(\boldsymbol{h}) \odot {\rm sin}S(\boldsymbol{\theta}) 
	\end{bmatrix} ,
	\nonumber
\end{align}
where $S$ reshapes all the $\boldsymbol{o}$, $\boldsymbol{h}$, and $\boldsymbol{\theta}$ to the size of $C'H_sW_s\times1$, the size of both $\boldsymbol{W}^{cos}$ and $\boldsymbol{W}^{sin}$ is ${C'H_sW_s}\times {C'H_sW_s}$, and $\odot$ represents element-wise multiplication.
According to the deduction outlined above, the PATM mechanism in the Wave-MLP block enables the fusion of all tokens within the feature maps using a single trainable parameter matrix. 
In most Siamese tracking settings, the number of tokens $H_sW_s$ is relatively small and fixed (determined by the backbone’s stride \cite{zhang2019deeper}). Thus, this matrix has a size of $2C'H_sW_s \times C'H_sW_s$, which implies that the parameters and FLOPs of a Wave-MLP block exhibit quadratic growth as the channel number $C'$ increases.

\subsection{Similarity Measurement of CFM}
\label{append:similarity_measurement_of_cfm}
CFM (the pre-fusion module in MCAS supernet) fuses template feature maps and search feature maps in a attention-like way implicitly if deducing the formula below,
\begin{align}
	\boldsymbol{y}& ={\rm Mutmal}(S(\boldsymbol{x}), S(\boldsymbol{F}^t)) = {\rm Mutmal}(S(\boldsymbol{r}{+_b}\boldsymbol{F}^s), S(\boldsymbol{F}^t)) \nonumber \\ 
	& = 
	\begin{bmatrix}        
		F^s_{1,1}+r_1 & 
		F^s_{1,2}+r_2 & 
		\cdots & 
		F^s_{1,C}+r_C \\        
		F^s_{2,1}+r_1 & 
		F^s_{2,2}+r_2 & 
		\cdots & 
		F^s_{2,C}+r_C \\       
		\vdots       & 
		\vdots       & 
		\ddots & 
		\vdots       \\        
		F^s_{H_sW_s,1}+r_1 & 
		F^s_{H_sW_s,2}+r_2 & 
		\cdots & 
		F^s_{H_sW_s,C}+r_C \\        
	\end{bmatrix} 
	\begin{bmatrix}        
		F^t_{1,1} & 
		F^t_{1,2} & 
		\cdots & 
		F^t_{1,H_tW_t} \\        
		F^t_{2,1} & 
		F^t_{2,2} & 
		\cdots & 
		F^t_{2,H_tW_t} \\       
		\vdots   & 
		\vdots   & 
		\ddots & 
		\vdots   \\        
		F^t_{C,1} & 
		F^t_{C,2} & 
		\cdots & 
		F^t_{C,H_tW_t} \\       
	\end{bmatrix},
	\nonumber
\end{align}
where $C$ is channel number of template and search feature maps, $S$ reshapes $\boldsymbol{r}{+_b}\boldsymbol{F}^s$ and $\boldsymbol{F}^t$ to the size of $H_sW_s\times C$ and $C\times H_tW_t$, $\boldsymbol{y}\in {\mathbb R}^{H_sW_s\times H_tW_t}$. 
The element in row $\phi$ and column $\varphi$ of $\boldsymbol{y}$ is,
\begin{equation} 
	y_{\phi\varphi} = \sum_{\psi=1}^C 
	{(F^s_{\phi\psi}+r_\psi)F^t_{\psi\varphi}}
	= \sum_{\psi=1}^C {F^s_{\phi\psi}F^t_{\psi\varphi}} 
	+ \sum_{\psi=1}^C {r_\psi F^t_{\psi\varphi}}.
	\label{eq_attention}
\end{equation}

The former term of Equation~\eqref{eq_attention} computes the similarity of template feature maps and search feature maps at the pixel level, which corresponds to the cross-attention mechanism of the CFA module in TransT \cite{chen2021transformer}. The latter term of Equation~\eqref{eq_attention} matches the features of the object region with the feature maps of the template. In this way, it enhances the representation of the object region and weakens the representation of the background region comparatively, corresponding to the self-attention mechanism of the ECA module in TransT.

Since the size of $\boldsymbol{y}$ can be reshaped to $H_s\times W_s\times C$, a fully connected layer can be applied to transform $\boldsymbol{y}$ into $\boldsymbol{z}$, whose size is $H_s\times W_s\times C'$. Here, $C'$ (typically 256) represents the channel number required by the subsequent Wave-MLP blocks. This transformation is mathematically represented as $\boldsymbol{z} = \boldsymbol{y}\boldsymbol{W} + \boldsymbol{b}$, where $\boldsymbol{W}\in{\mathbb R}^{H_tW_t\times C'}$ and $\boldsymbol{b}\in{\mathbb R}^{H_sW_s\times C'}$ are trainable parameters.
The element in row $\phi$ and column $\zeta$ of $\boldsymbol{z}$ is given by:
\begin{equation}
	\begin{split}
		z_{\phi\zeta} 
		= 
		\sum_{\varphi=1}^{H_tW_t} 
		{y_{\phi\varphi}w_{\varphi\zeta}}+b_{\phi\zeta}
		&= 
		\sum_{\varphi=1}^{H_tW_t}
		{
			\left(
			\sum_{\psi=1}^C {F^s_{\phi\psi}F^t_{\psi\varphi}} 
		    + 
		    \sum_{\psi=1}^C {r_\psi F^t_{\psi\varphi}}
		    \right)
		w_{\varphi\zeta}
		}
		+
		b_{\phi\zeta} 
		\\ &= 
		\sum_{\psi=1}^C 
		{
			F^s_{\phi\psi}
			\left(
			\sum_{\varphi=1}^{H_tW_t} {F^t_{\psi\varphi}w_{\varphi\zeta}}
			\right)
		} 
		+ 
		\sum_{\psi=1}^C 
		{
			r_\psi
			\left(
			\sum_{\varphi=1}^{H_tW_t} {F^t_{\psi\varphi}w_{\varphi\zeta}}
			\right)
		} 
		+ b_{\phi\zeta},
	\end{split}
	\nonumber
\end{equation}
where $F^t_{\psi\varphi}w_{\varphi\zeta}$ indicates that the linear transformation is applied to the template feature maps. With these trainable parameters, the similarity measurement becomes more representative and flexible, avoiding information loss from the search region. The output $\boldsymbol{z}$ of CFM contains more differentiated feature tokens, allowing for better establishment of global dependencies by the Wave-MLP to address specific challenges in tracking.

\subsection{Inverse Chained Cost Estimation}
\label{append:inverse_chained_cost_estimation}
Referring to \cite{fang2020densely}, our computational cost (latency or FLOPs measurement) is optimized using a inverse chained cost estimation algorithm. This algorithm considers the global effects of connections on the predicted cost when adopting our MCAS-relaxed supernet. We measure the cost of each Wave-MLP block \cite{tang2022image} separately in advance and record them in a lookup table. The cost of one basic layer is estimated as follows:
\begin{equation}
	{\rm cost}_{\text{CH/PH}}^{i,l} = 
	\sum_{o\in \mathcal{O}^c} {
		\frac{{\rm exp}({\alpha}_o^{i,l})}{
			\sum_{o'\in \mathcal{O}^c}{{\rm exp}({\alpha}_{o'}^{i,l})}}
	\cdot {\rm cost}_o^{i,l}},
	\nonumber
\end{equation}
where ${\rm cost}_o^{i,l}$ represents the pre-measured cost of the Wave-MLP block $o\in \mathcal{O}^c$ in the basic layer $l$ within the harmonization block $H_i$. ${\rm cost}_{\text{CH}}^{i,l}$ refers to the $l$-th basic layer in the Channel-Harmonization layers, while ${\rm cost}_{\text{PH}}^{i,l}$ denotes the $l$-th basic layer in the Pixel-Harmonization layers.

The chained cost estimation algorithm is executed as follows:
\begin{equation}
	\hat{{\rm cost}}^{i} ={\rm cost}_{\text{PH}}^i + \sum_{j=i+1}^{i+m'} {
		\frac{{\rm exp} (\beta_{i,j})}
		{\sum_{j'=1}^{m'} {{\rm exp} (\beta_{i,j'})}} 
		\cdot ({\rm cost}_{\text{CH}}^{ij} + {\rm cost}_{\text{PH}}^{j})},
	\nonumber
\end{equation}
where ${\rm cost}_{\text{PH}}^i$ represents the total cost of all the basic fusion layers of $H_i$, which can be computed as a summation, ${\rm cost}_{\text{PH}}^i=\sum_{l}{{\rm cost}_{\text{PH}}^{i,l}}$. The numbers of preceding/subsequent harmonization blocks connected to $H_i$ are denoted by $m$ and $m'$. ${\rm cost}_{\text{CH}}^{ij}$ denotes the cost of the basic layer in the Channel-Harmonization layers in $H_j$, which processes the channel number of the feature map from $H_i$.

The overall cost of our supernet can be obtained by computing $\hat{\rm cost}^1$ recursively, i.e., ${\mathcal{L}}_{sea}= \hat{{\rm cost}}^1$.

\subsection{Complexity Analysis}
\label{append:complexity_analysis}
There is a comprehensive comparison of the \textit{computational complexity} (measured in FLOPs) for the key modules in our fusion pipeline: the Coarse Fusion MLP (CFM) and Refine Fusion MLP (RFM). The results are summarized in Table~\ref{tab:flops_all_blocks}, where we compare our designs with commonly used alternatives in visual tracking literature.
\begin{table}[h]
	\caption{FLOPs comparison of different fusion blocks used in CFM and RFM}
	\centering
	\label{tab:flops_all_blocks}
	\resizebox{0.72\linewidth}{!} 
	{
		\begin{tabular}{ccc}
			\hline
			\textbf{Module} & \textbf{Block Type} & \textbf{FLOPs Expression} \\ \hline
			\multirow{3}{*}{CFM} 
			& Convolution-based & $H_s W_s \cdot H_t W_t \cdot C \cdot C'$ \\
			& Cross-attention-based & $2H_s W_s \cdot H_t W_t \cdot C + H_s W_s \cdot C \cdot C'$ \\
			& \textbf{CFM (Ours)} & $H_s W_s \cdot H_t W_t \cdot (C + C') + H_t W_t \cdot C$ \\ \hline
			\multirow{4}{*}{RFM}
			& Standard CNN (3$\times$3) & $9 \cdot H_z W_z \cdot C \cdot C'$ \\
			& Depthwise CNN & $H_z W_z \cdot (9C + C \cdot C')$ \\
			& Transformer (4 heads, w/o FFN) & $4 \cdot (H_z W_z)^2 \cdot C + 4 \cdot H_z W_z \cdot C \cdot C'$ \\
			& \textbf{Wave-MLP (used in our method)} & $2 \cdot (H_z W_z)^2 \cdot C + 2 \cdot H_z W_z \cdot C \cdot C'$ \\ \hline
		\end{tabular}
	}
	\begin{flushleft}
		\footnotesize
		*$H_s{\times}W_s$ and $H_t{\times}W_t$ denote spatial sizes of search feature $\boldsymbol{F}^s$ and template feature $\boldsymbol{F}^t$ respectively, while $H_z{\times}W_z$ represents the coarsely fused feature $\boldsymbol{z}$ from CFM; $C$ and $C'$ indicate input/output channels (constant terms like bias are omitted).
	\end{flushleft}
\end{table}

\paragraph{CFM Module}
This module fuses information from both the template and search branches. We compare three designs:
\begin{itemize}
	\item \textbf{Convolution-based fusion:} This approach treats the template feature map as convolutional kernels and slides it over the search feature. It is efficient (lowest FLOPs), but it only enables \textit{patch-level similarity} and lacks fine-grained pixel-wise alignment \cite{zhang2021learn}.
	
	\item \textbf{Cross-attention-based fusion:} This enables pixel-level interactions via full attention matrices. However, the complexity grows significantly with the product of template and search spatial dimensions, and it requires multiple projections and softmax operations.
	
	\item \textbf{Our CFM:} We propose a lightweight alternative that uses ROI-aligned template features and linear fusion. The complexity is similar to attention-based methods, but CFM avoids softmax and multi-head operations, making it significantly faster in practice and more hardware-friendly. Its structure allows pixel-level fusion while maintaining low computational overhead. The last term $H_t W_t \cdot C$ results from pooling template features before fusion, which introduces negligible cost.
\end{itemize}

\paragraph{RFM Module}
This module operates on a single fused feature map $\boldsymbol{z}$ and enhances spatial and channel-wise representations. Four variants are considered:
\begin{itemize}
	\item \textbf{Standard 3$\times$3 CNN:} Efficient but limited in capturing global dependencies due to small receptive fields.
	
	\item \textbf{Depthwise CNN:} Reduces parameter count by decoupling spatial and channel mixing, but still lacks global context modeling.
	
	\item \textbf{Transformer:} Offers global receptive field and pixel-level interaction, but incurs \textbf{quadratic FLOPs} due to its all-to-all attention ($O((HW)^2)$), making it impractical for high-resolution feature maps.
	
	\item \textbf{Wave-MLP (adopted in our method):} Introduced in \cite{tang2022image}, Wave-MLP achieves global token mixing via phase-aware transformations with lower complexity than Transformers. As shown in Table~\ref{tab:flops_all_blocks}, although the first term scales quadratically with spatial size, the constant is significantly smaller than that of Transformers (2 vs. 4 heads), and the absence of softmax, multi-head projections, and attention maps makes it faster and more parallelizable on edge devices.
\end{itemize}

\section{More Details of Experiment}
\label{append:more_details_of_experiment}

\subsection{Overall Performance}
\label{append:overall_performance}
\begin{figure}[h]
	\centering
	\includegraphics[width=\linewidth]{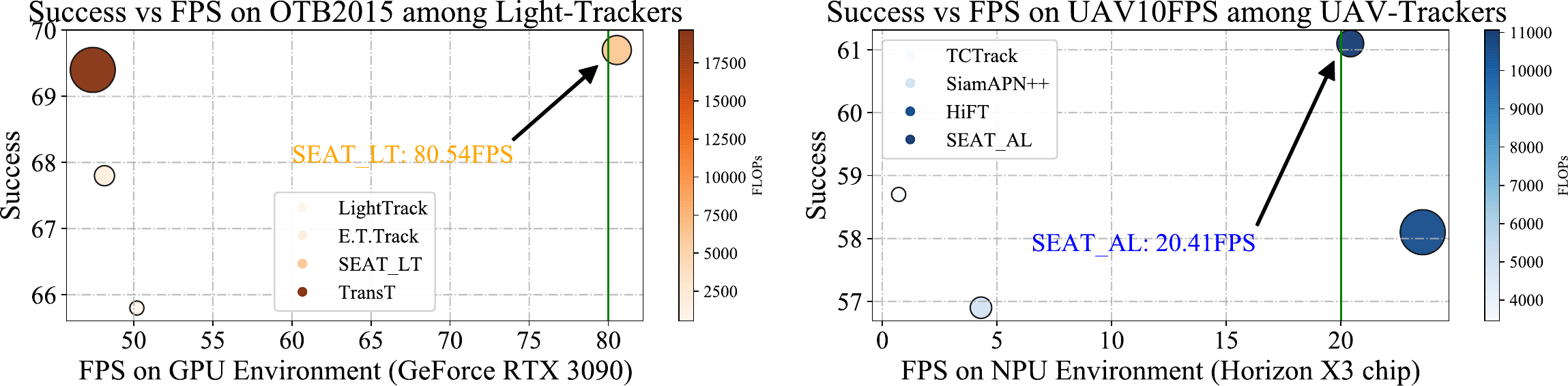}
	\caption{
        Performance comparison of trackers in terms of frames per second (FPS) and Success Rate. The visualizations represent computational complexity (FLOPs) by color intensity and model size by circle size. The green line marks the real-time FPS threshold. 
	}
	\label{fig_fvs}
\end{figure}
The overall performance of SEAT\_LT and SEAT\_AL is presented in Figure~\ref{fig_fvs}, demonstrating an effective balance between accuracy and efficiency. SEAT attains top-tier accuracy on four general-purpose and three aerial tracking benchmarks, while delivering superior speed on both a GeForce RTX 3090 GPU and a Horizon X3 NPU edge chip.

\subsection{Attribute Analysis of UAV Trackers}
\label{append:attribute_analysis_of_uav_trackers}
The performance of SEAT\_AL was thoroughly evaluated across 12 aerial-specific attributes, as outlined by \cite{mueller2016benchmark}. These attributes represent common challenges in aerial tracking: IV (Illumination Variation), LR (Low Resolution), FM (Fast Motion), CM (Camera Motion), SOB (Similar Object), ARC (Aspect Ratio Change), FOC (Full Occlusion), SV (Scale Variation), OV (Out-of-View), VC (Viewpoint Change), BC (Background Clutters), and POC (Partial Occlusion).
\begin{figure}[h]
	\centering
	\includegraphics[width=10cm]{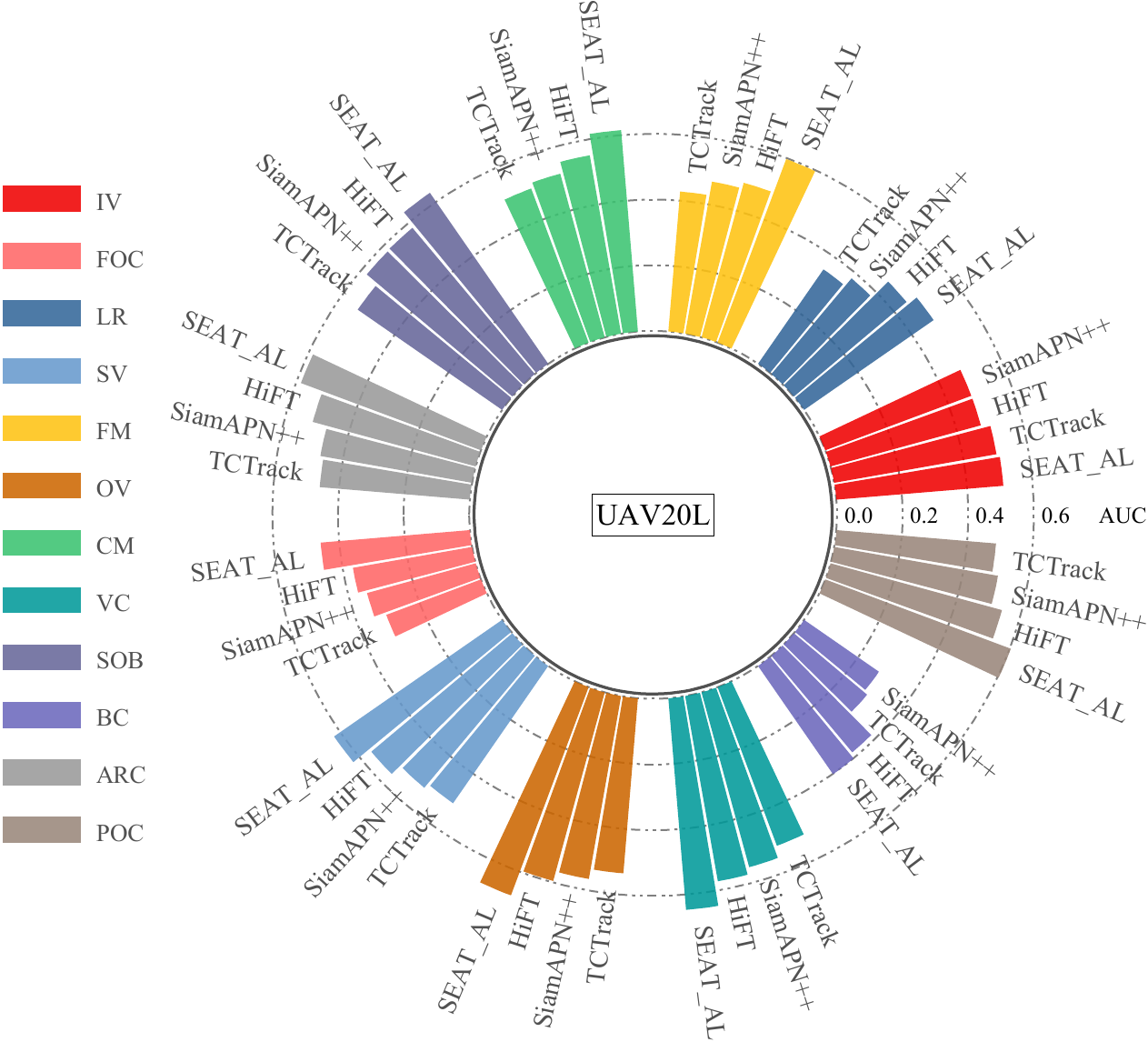}
	\caption{Qualitative Attribute-based comparisons of trackers for NPU deployment on UAV20L. The top five trackers are displayed in a bar chart, illustrating their success scores across 12 different attributes, with each bar representing a specific attribute.}
	\label{fig_attr_al}
\end{figure}

As shown in Figure~\ref{fig_attr_al}, SEAT\_AL demonstrates strong performance in handling these challenges, particularly excelling in Camera Motion (CM) and Viewpoint Change (VC) — scenarios that are especially difficult in aerial tracking due to the dynamic nature of drones. These high scores suggest that SEAT\_AL is well-suited to managing the complexities of aerial footage, such as rapid camera movements and significant changes in the tracked object's appearance.

The success of SEAT\_AL can be attributed to its use of MLP modules, which effectively capture global dependencies and pixel relationships essential in handling fast and complex motion. This allows SEAT\_AL to maintain tracking robustness even during challenging flight maneuvers and viewpoint shifts.
Overall, the attribute analysis highlights the versatility and robustness of SEAT\_AL, affirming its capability as a reliable tracker for aerial tracking environments that require NPU on-board deployment.

\subsection{Search Space and Derived Model}
\label{append:search_space_and_derived_model}
\begin{figure*}[h]
	\centering
	\includegraphics[width=\linewidth]{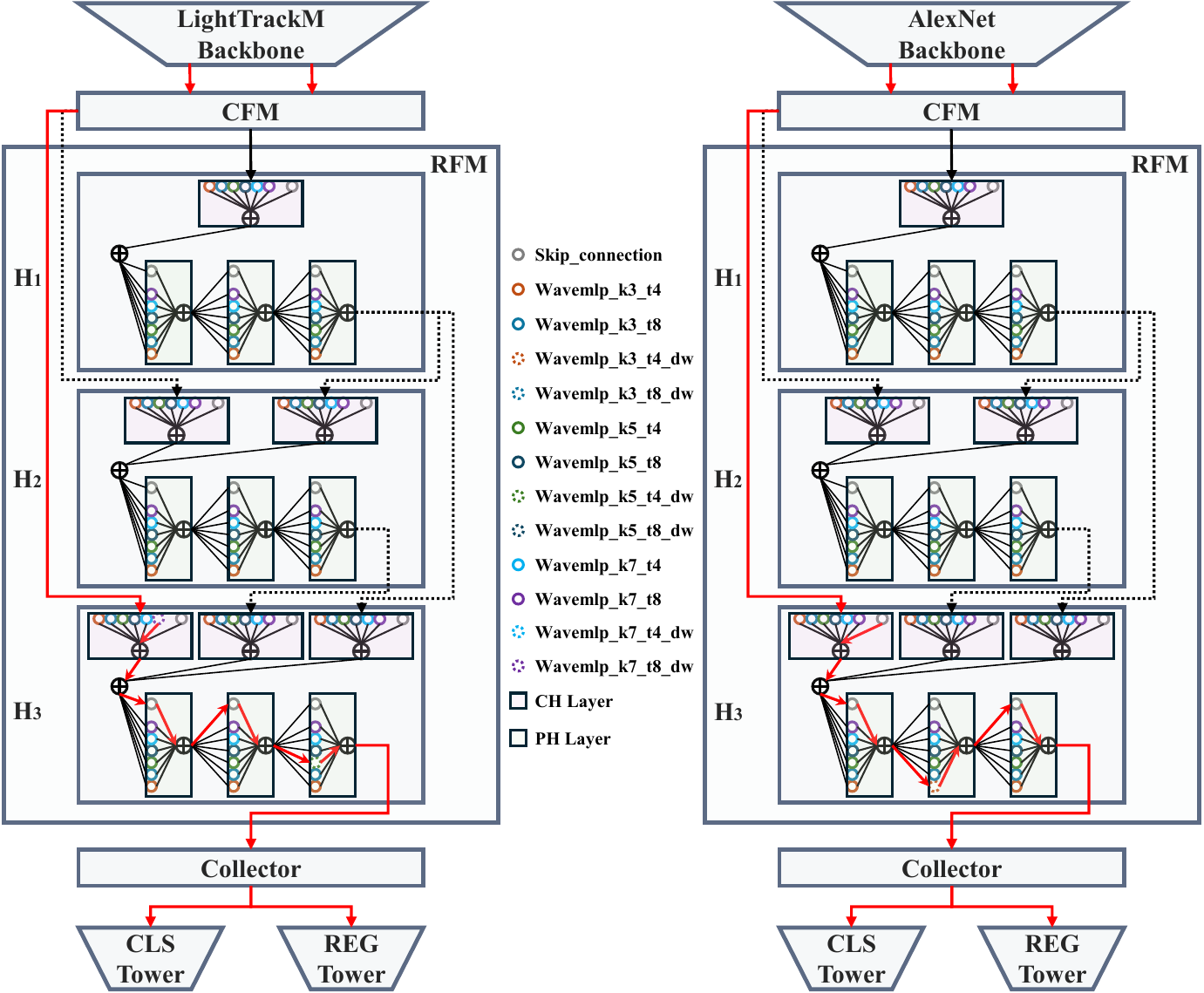}
	\caption{
        Visualization of the MCAS-relaxed supernet and derived architectures for resource-constrained targets (SEAT\_LT for GPU, SEAT\_AL for NPU).
	}
	\label{fig_search}
\end{figure*}

As shown in Figure~\ref{fig_search}, the complete search space for the correlation operation in the neck of our Siamese tracker comprises a fixed Coarse Fusion MLP (CFM) Module and a Refine Fusion MLP (RFM) Module, which varies across different search iterations. In our experiments, the RFM consists of three Harmonization Blocks, which are densely connected. Specifically, the first Harmonization Block only receives the feature map from the CFM as input, the second block receives both the CFM input and the output from the previous block, and this pattern continues for subsequent blocks.

The feature maps produced by the three Harmonization Blocks are unified to a consistent number of channels, specifically 256, 320, and 384 in each block, respectively. The task of unifying the channels is handled by the Channel Harmonization (CH) layers within each Harmonization Block, while the Pixel Harmonization (PH) layers continue to enhance and fuse the features at the pixel level. Each CH and PH layer is a Basic Layer, which is composed of different configurations of Wave-MLP blocks. For example, a configuration labeled as WaveMLP\_k5\_t4\_dw specifies a Wave-MLP block with a kernel size of 5, an MLP expansion ratio of 4, and depth-wise convolution. The outputs of these blocks are combined point-wise and passed to subsequent layers. In the end, each Basic Layer selects only one Wave-MLP block for the final architecture.

As shown in Figure~\ref{fig:pipeline} of the main text, we configured the supernet, composed of this search space, with different backbone architectures. After the search phase, we derived two models from the supernet: SEAT\_LT for resource-constrained GPU usage, and SEAT\_AL for resource-limited NPU platforms. The specific paths used by these two MCAS-based trackers in the supernet are highlighted in red in Figure~\ref{fig_search}.

\subsection{Ablation of NAS Approach}
\label{append:ablation_of_nas_approach}
To demonstrate the superiority of our differentiable neural architecture search (DNAS) method based on the Harmoni-relaxation strategy, we compare it with evolutionary neural architecture search (ENAS), which has been employed in \cite{yan2021lighttrack} to search for lightweight backbones in Siamese tracking. Specifically, we maintain the Wave-MLP blocks in the search space but use an evolutionary algorithm to continuously search for the optimal block configurations and combinations.

\begin{figure}[h]
	\centering
	\includegraphics[width=\linewidth]{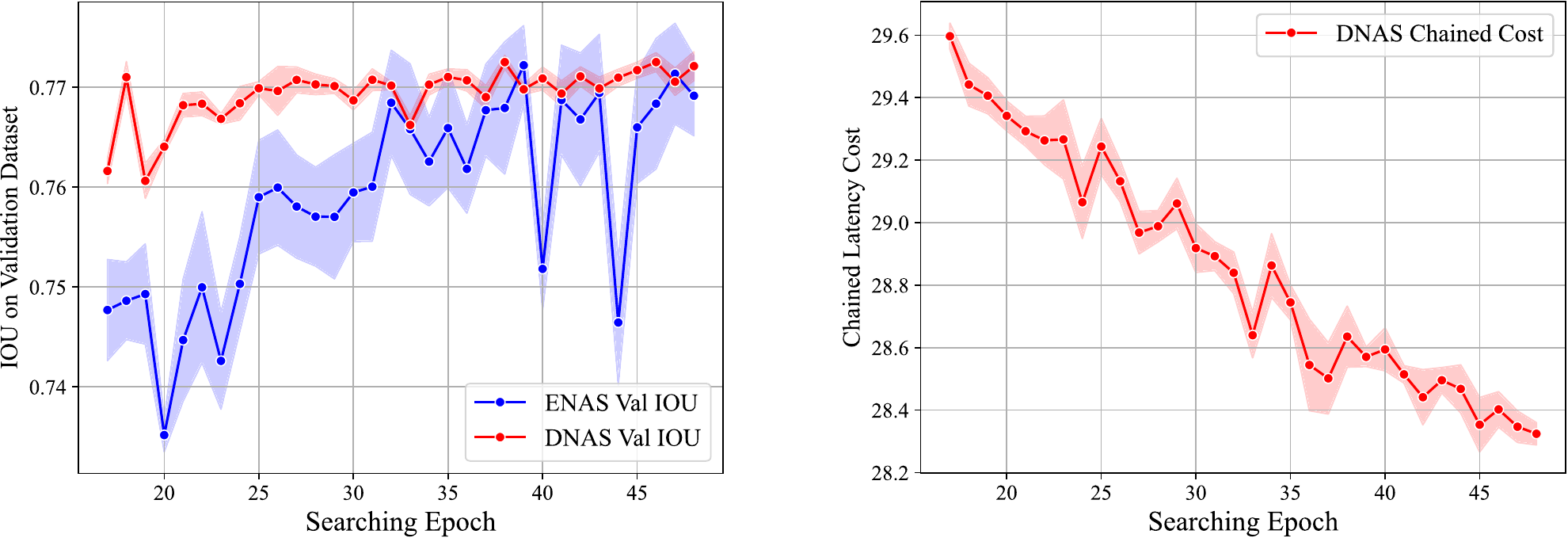}
	\caption{Comparison of the search processes between differentiable neural architecture search (DNAS) and evolutionary neural architecture search (ENAS). The x-axis denotes the number of search iterations, while the y-axis represents the Intersection over Union (IOU) metric on the validation set (left) and the total chained latency cost (right).}
	\label{fig_nas}
\end{figure}
As illustrated in Figure~\ref{fig_nas}, our algorithm, which relaxes the search space into a continuous space, achieves a more stable search process with smaller variance. This continuous space includes more detailed candidate possibilities, resulting in higher accuracy on the validation set. In contrast, the evolutionary algorithm introduces mutations, leading to greater uncertainty in the search process and consequently larger variance. The discrete space search of ENAS results in slightly lower accuracy on the validation set. However, it is worth noting that the search time overhead of DNAS is significantly higher than that of ENAS.
\end{document}